\pdfoutput=1
\documentclass[5p,times]{elsarticle}
\usepackage[english]{babel}
\usepackage{amsmath}
\usepackage{amssymb}
\usepackage{graphicx}
\usepackage{tabularx}
\usepackage{array}
\usepackage{booktabs}
\usepackage{makecell}
\usepackage{algorithm}
\usepackage{algpseudocode}
\usepackage{xspace}
\usepackage{xcolor}
\usepackage{microtype}
\usepackage[colorlinks=true, allcolors=blue]{hyperref}
\hypersetup{
    pdftitle={Context-Aware Error Mitigation Orchestration for Hybrid Quantum Reinforcement Learning on NISQ Systems},
    pdfauthor={Bisma Majid, Shabir Ahmed Sofi, Mir Mohammad Yousuf},
    pdfkeywords={Quantum reinforcement learning, Error mitigation, NISQ, Vehicle routing problem}
}

\algnewcommand\algorithmicforeach{\textbf{for each}}
\algdef{S}[FOR]{ForEach}[1]{\algorithmicforeach\ #1\ \algorithmicdo}

\makeatletter
\def\ps@pprintTitle{%
  \let\@oddhead\@empty
  \let\@evenhead\@empty
  \let\@oddfoot\@empty
  \let\@evenfoot\@oddfoot}
\makeatother

\begin{document}

\begin{frontmatter}

\title{Context-Aware Error Mitigation Orchestration for Hybrid Quantum Reinforcement Learning on NISQ Systems}

\author[nits]{Bisma Majid}
\ead{bismabhat_ite006@nitsri.ac.in}
\author[nits]{Shabir Ahmed Sofi}
\ead{shabir@nitsri.ac.in}
\author[nits]{Mir Mohammad Yousuf\corref{cor1}}
\ead{yousuf_2022phaite006@nitsri.ac.in}

\cortext[cor1]{Corresponding author.}

\affiliation[nits]{organization={Department of Information Technology, National Institute of Technology Srinagar},
addressline={Hazratbal}, city={Srinagar}, postcode={190006}, state={Jammu and Kashmir}, country={India}}

\begin{abstract}
Quantum Reinforcement Learning (QRL) integrates reinforcement learning with parameterized quantum circuits and is a promising approach to combinatorial optimization. On Noisy Intermediate-Scale Quantum (NISQ) devices, however, decoherence, gate imperfections, and measurement errors reduce policy quality and make learning less reliable. Existing error mitigation techniques are generally applied as fixed corrections that do not adapt to changing noise conditions or to the evolving state of training. This work presents Adaptive Policy-Guided Error Mitigation (APGEM) as a context-aware orchestration layer of the hybrid quantum-classical training loop that dynamically selects the most suitable mitigation strategy during QRL training. APGEM evaluates Zero-Noise Extrapolation (ZNE), Probabilistic Error Cancellation (PEC), Clifford Data Regression (CDR), and Readout Error Mitigation (REM) using policy-level indicators, including quantum-state fidelity, policy entropy, cumulative reward, and approximation ratio, and integrates the selected strategy directly into the reinforcement learning loop. The framework is evaluated on the Capacitated Vehicle Routing Problem (CVRP), a representative NP-hard problem in urban logistics, under a range of NISQ noise models and noise levels. APGEM consistently outperforms conventional static mitigation methods, reaches approximately 94\% of the utility of an oracle strategy, maintains higher quantum-state fidelity as noise increases, and produces more stable learning behaviour throughout training. Ablation studies show that the framework learns context-aware mitigation policies that adapt to different noise environments and circuit execution conditions. These findings demonstrate that integrating adaptive error mitigation into the learning process substantially improves the robustness and reliability of QRL on NISQ hardware.
\end{abstract}


\end{frontmatter}

\section{Introduction}

Combinatorial optimization problems such as the Vehicle Routing Problem (VRP) represent some of the most challenging and practically relevant tasks in operations research and logistics \cite{vidal2022hybrid,mor2022vehicle}. The VRP, in its many variants, requires the design of cost-efficient routes for a fleet of vehicles to serve a set of customers under resource and operational constraints such as vehicle capacity, service time, and time windows \cite{ambrosino2022rich}. Efficient solutions to VRPs are critical for applications in transportation networks, supply chain management, postal and parcel delivery, and distribution systems, where even marginal reductions in routing cost can yield significant economic savings, reduced fuel consumption, and lower environmental impact \cite{sabet2022green,los2022large}. Given the scale and complexity of modern logistics systems, VRPs are not only of theoretical importance but also a key driver of efficiency in industry and sustainability initiatives \cite{fakhravar2022combining}.

Classical approaches to VRPs include exact solvers based on mixed-integer programming and a wide range of heuristics and metaheuristics, such as nearest-neighbor, tabu search, simulated annealing, ant colony optimization, and genetic algorithms \cite{ochelska2021selected}. While these methods remain powerful and are widely applied in practice, they often struggle to scale effectively with problem size. The VRP is NP-hard, and as the number of customers, vehicles, and constraints increases, the computational burden grows exponentially \cite{ghosal2024unifying}. Heuristic methods, though effective for moderate scales, tend to degrade in performance for large-scale, high-dimensional instances with complex temporal and capacity constraints \cite{bouanane2022vehicle}. This motivates the search for novel computational paradigms that can complement classical methods in tackling large-scale optimization.

With the emergence of noisy intermediate-scale quantum (NISQ) devices, quantum computing has been identified as a promising candidate for advancing combinatorial optimization \cite{preskill2023quantum,bharti2022noisy}. Quantum algorithms, by leveraging superposition and entanglement, have the potential to explore vast solution spaces in ways that classical methods cannot. While large-scale, fault-tolerant quantum computers are still beyond reach, hybrid quantum-classical algorithms designed for NISQ-era devices have become central to ongoing research \cite{lamichhane2025quantum,majid2025quantum,yousuf2026systematic}. Among these, Variational Quantum Algorithms (VQAs) are especially promising due to their compatibility with near-term devices, reliance on shallow circuits, and adaptability through classical feedback loops \cite{cerezo2021variational}.

Quantum Reinforcement Learning (QRL) extends this paradigm by embedding parameterized quantum circuits within reinforcement learning agents, enabling policies to be represented as variational quantum circuits (VQCs) \cite{kolle2024study}. In this setting, the quantum policy maps environment states to action probabilities, while classical optimization updates the parameters. This hybrid structure makes QRL particularly attractive for dynamic decision-making problems such as VRP, where sequential adaptability and exploration of combinatorial action spaces are essential. However, QRL faces two central challenges in practice. First, quantum circuits executed on NISQ devices are highly susceptible to noise, which degrades fidelity, increases state entropy, and ultimately destabilizes the learning process \cite{kim2023scalable}. Second, while error mitigation strategies have been developed to counteract noise, they are typically applied in a static manner. Conventional mitigation methods such as Zero-Noise Extrapolation (ZNE), Probabilistic Error Cancellation (PEC), Clifford Data Regression (CDR), and Readout Error Mitigation (REM) are generally selected once and applied uniformly, without adapting to the varying characteristics of circuits or noise across different executions \cite{pelofske2025digital}. This static application often results in suboptimal improvements and fails to fully exploit the potential of mitigation in a reinforcement learning context.

Recent progress in error mitigation has shown that techniques such as Zero-Noise Extrapolation (ZNE), Probabilistic Error Cancellation (PEC), Clifford Data Regression (CDR), and Readout Error Mitigation (REM) can substantially improve quantum execution quality when carefully applied \cite{nation2021scalable,ding2022evaluating}. However, existing studies generally evaluate these techniques in isolation and apply them as fixed mitigation strategies throughout execution. Such static approaches fail to account for the dynamic nature of NISQ noise, where the effectiveness of a mitigation technique may vary with circuit characteristics, noise intensity, and the current stage of learning. Consequently, there remains an open challenge of designing adaptive mitigation frameworks that can continuously identify and apply the most suitable mitigation strategy based on real-time performance signals \cite{yousuf2026systematic}. Addressing this challenge is particularly important in Quantum Reinforcement Learning (QRL), where noise directly affects policy optimization and learning stability \cite{czarnik2021error}.

In this work, we extend the \textbf{Adaptive Policy-Guided Error Mitigation (APGEM)} framework, first proposed for the Traveling Salesman Problem by Majid et al.~\cite{majid2025noise}, to Quantum Reinforcement Learning applied to capacitated VRP instances. The framework combines reinforcement learning, variational quantum circuits, and adaptive error mitigation in a closed feedback loop. Our key contributions are fourfold:

\begin{enumerate}
    \item \textbf{Hybrid QRL Environment for VRP:} We design a reinforcement learning environment tailored to VRP, where the policy is represented by a variational quantum circuit trained under noisy simulations that mimic realistic NISQ hardware conditions. This provides a practical benchmark for evaluating quantum-enhanced learning.
    \item \textbf{APGEM Controller:} We introduce an adaptive error mitigation controller that dynamically selects among multiple mitigation techniques (ZNE, PEC, CDR, REM). The controller employs a LinUCB contextual-bandit selection strategy over a context vector (depolarizing noise level, readout rate, fidelity and its degradation, entropy, normalized circuit depth, remaining shot budget, approximation ratio, and recent reward trend), scored by a cost-aware utility that balances mitigated fidelity against sampling cost and estimator variance. This transforms error mitigation from a static pre-processing step into a dynamic, shot-budget-aware runtime component of the hybrid quantum-classical loop, integrated with reinforcement feedback.
    \item \textbf{Feasibility-Guaranteed Policy-Gradient Training:} We train the quantum policy with REINFORCE and exact parameter-shift gradients over a coverage-preserving decoder with cheapest-insertion repair, which guarantees complete feasible routes on every instance and provides a directed learning signal for the routing policy.
    \item \textbf{Comprehensive Adaptive Mitigation Analysis:} We conduct extensive ablation and robustness studies to analyze mitigation-selection behavior across varying noise regimes, circuit characteristics, and training horizons. The results reveal that APGEM learns interpretable and context-aware mitigation policies that closely approach oracle-level mitigation performance while consistently outperforming fixed mitigation strategies.
\end{enumerate}

Through extensive experiments across different noise families (depolarizing, amplitude damping, dephasing, and readout) and multiple noise levels, we demonstrate that APGEM consistently outperforms static mitigation strategies and approaches oracle-level mitigation performance. The framework learns context-dependent mitigation policies that adapt to changing operating regimes, preserves higher quantum-state fidelity under increasing noise, reduces performance degradation, and improves learning stability throughout training. These findings highlight the importance of adaptive error mitigation as a key enabling technology for reliable Quantum Reinforcement Learning on NISQ devices and demonstrate its potential for real-world combinatorial optimization applications.

The remainder of this paper is organized as follows. Section~\ref{sec:related} reviews the related work and highlights the existing approaches and gaps in the literature. Section~\ref{sec:problem} presents the problem formulation, where the key challenges and objectives are formally defined. Section~\ref{sec:methodology} details the proposed methodology, including the underlying framework and techniques employed. Section~\ref{sec:metrics} introduces the performance metrics that are used to assess the effectiveness of the proposed approach. Section~\ref{sec:setup} describes the experimental setup, including datasets, noise models, and parameter settings. Section~\ref{sec:results} reports and analyzes the results obtained, with comparisons across different scenarios. Finally, Section~\ref{sec:conclusion} concludes the paper with key findings, limitations, and possible directions for future research.

\section{Related Work}\label{sec:related}

Quantum computing in logistics and supply chain management has been widely discussed in recent overviews such as \cite{1phillipson2024quantum} and \cite{6perez2024quantum}, which emphasize the suitability of problems like TSP, VRP, and network flow for quantum optimization. These works highlight the promise of quantum algorithms, particularly QAOA and annealing, but also note the challenges posed by NISQ-era devices, including qubit limitations, noise, and shallow depth constraints. The research gap here lies in the lack of strategies that directly address the noise-resilience of logistics-focused quantum algorithms. Our work differs by not only applying quantum reinforcement learning (QRL) to CVRP but also embedding an adaptive error mitigation layer that explicitly counters NISQ noise during learning.

In parallel, studies on quantum-inspired computing such as \cite{2nunez2024quantum} and \cite{8whig2024quantum} demonstrate that classical algorithms inspired by quantum principles can yield scalable heuristics and practical benefits for logistics planning. While these approaches serve as useful baselines, they cannot capture the full advantages of true quantum superposition and interference. Moreover, they lack mechanisms to adapt error management dynamically, since they are executed on classical systems. In contrast, our APGEM-QRL framework leverages actual quantum circuits simulated under noise and adapts mitigation strength in real time, going beyond static quantum-inspired methods.

Research focusing on direct quantum algorithms for logistics optimization (for example, \cite{4sonavaneexploring}) shows how logistics problems can be mapped into Ising or QUBO formulations. These studies explore penalty-based encodings, circuit depth trade-offs, and hybrid decomposition strategies, but they generally assume fixed noise levels or apply static mitigation techniques. The gap here is the absence of noise-aware learning that evolves with circuit complexity and training phases. Our approach addresses this by coupling reinforcement learning with Adaptive Policy-Guided Error Mitigation (APGEM), allowing the quantum solver to adaptively regulate noise handling based on policy progress and circuit depth.

Cognitive and learning-oriented perspectives such as \cite{3acuna2025quantum} introduce the notion of quantum-enhanced cognitive modeling for logistics, suggesting that quantum interference can improve policy exploration. While these studies provide a conceptual foundation for quantum reinforcement learning, they remain largely at the modeling stage and do not address the critical challenge of noise robustness. Our work extends this idea by not only using QRL for policy learning but also embedding mitigation strategies that respond to real-time feedback signals, thereby improving convergence reliability in noisy environments.

At the systems level, \cite{5singh2025quantum} and \cite{7singh2026quantum} argue for the integration of quantum methods into end-to-end logistics workflows, highlighting real-world requirements such as data acquisition, hybrid orchestration, and translation of algorithmic outcomes into operational KPIs like cost, energy efficiency, and on-time delivery. These contributions provide a systems perspective but stop short of demonstrating noise-resilient learning with statistical validation. Our work addresses this by evaluating on standard CVRPLIB benchmark instances under multiple noise models, and reporting not just raw optimization cost but also approximation ratio to best-known solutions, reward convergence, distribution fidelity, and entropy, together with formal statistical testing across instances.

Finally, \cite{8whig2024quantum} emphasizes the importance of data-driven and adaptive decision-making in logistics optimization. While this line of work showcases the benefits of feedback mechanisms in classical settings, it does not extend adaptivity to quantum error mitigation. Here lies another gap: existing approaches do not treat error mitigation as a controllable, feedback-driven process. APGEM fills this gap by making mitigation strength a dynamic variable guided by policy performance, noise conditions, and learning stage, transforming mitigation from a static parameter into an adaptive component of the training loop.

APGEM was first proposed for QRL on the Traveling Salesman Problem by Majid et al.~\cite{majid2025noise}, where adaptively selecting among mitigation techniques improved noise resilience over static mitigation, and was subsequently applied with an $\varepsilon$-greedy selector to a real-world CVRP case study~\cite{sofi2026apgemcvrp}. The present study extends this framework to the capacity-constrained CVRP and contributes a LinUCB contextual-bandit controller with a cost-aware utility, a feasibility-guaranteed decoder, an expected-utility oracle and regret decomposition, and statistical validation on thirty CVRPLIB instances.

Taken together, prior research establishes the foundations for applying quantum and quantum-inspired methods to logistics but leaves several gaps unaddressed: static treatment of noise, limited focus on real-world urban networks, lack of integration between learning dynamics and error mitigation, and reliance on narrow performance metrics. This paper addresses these shortcomings by proposing APGEM-QRL, a hybrid framework that adaptively mitigates noise during QRL training on standard CVRPLIB benchmark instances, achieving improved approximation ratios, faster and more stable convergence, and enhanced robustness under varying noise conditions.

Table~\ref{tab:litreview} summarizes the focus of prior works, the gaps they leave unaddressed, and how our proposed APGEM-QRL framework advances the state of the art. As shown, while existing studies have explored quantum algorithms, quantum-inspired heuristics, and system-level integration for logistics, they generally treat error mitigation as static, rely on simplified problem encodings, or lack real-world urban validation. In contrast, our work introduces an adaptive, policy-guided error mitigation mechanism embedded directly within a quantum reinforcement learning framework, validated on realistic urban road graphs under diverse noise models.

\begin{table*}[!t]
    \centering
    \caption{Summary of prior works, identified gaps, and contributions of our Work.}
    \label{tab:litreview}
    \footnotesize
    \renewcommand{\arraystretch}{1.15}
    \begin{tabularx}{\textwidth}{@{}p{1.6cm}XXX@{}}
    \toprule
    \textbf{Ref(s)} & \textbf{Focus of Prior Work} & \textbf{Identified Gap} & \textbf{Contribution of our Work} \\
    \midrule
    {\cite{1phillipson2024quantum}, \cite{6perez2024quantum}} & Surveys of quantum computing in logistics and supply chains; discuss QAOA, annealing, and hybrid methods. & Do not address noise-resilience or adaptive error mitigation in NISQ settings. & Propose \textbf{APGEM-QRL}, embedding adaptive noise mitigation within a reinforcement learning framework. \\
    {\cite{2nunez2024quantum}, \cite{8whig2024quantum}} & Quantum-inspired algorithms for logistics optimization; scalable heuristics on classical hardware. & Lack of true quantum implementation and no adaptive error management. & Implement \textbf{noisy quantum circuits} with \textbf{feedback-driven error mitigation} beyond static QIC baselines. \\
    {\cite{4sonavaneexploring}} & Application of QAOA/annealing to logistics problems (TSP, VRP) via QUBO/Ising formulations. & Assumes fixed noise or applies static mitigation; no policy-aware adaptivity. & Introduce \textbf{policy-guided mitigation} that adjusts dynamically to circuit depth, reward dynamics, and noise type. \\
    {\cite{3acuna2025quantum}} & Cognitive/learning-inspired quantum models for route optimization and decision-making. & Conceptual focus; lacks robustness analysis under noisy quantum hardware. & Extend to \textbf{QRL with adaptive mitigation}, ensuring reliable convergence under real-world noise. \\
    {\cite{5singh2025quantum}\cite{7singh2026quantum}} & Systems perspective on integrating quantum methods into logistics workflows and enterprise KPIs. & Limited demonstrations on realistic urban road networks; no dynamic noise control. & Evaluate on \textbf{standard CVRPLIB benchmark instances} under diverse noise models, reporting best-known-referenced metrics with statistical testing. \\
    {\cite{8whig2024quantum}} & Data-driven adaptive decision-making in logistics using quantum-inspired techniques. & Adaptivity applied only to classical decision processes, not to quantum error mitigation. & Make \textbf{error mitigation adaptive}, guided by policy feedback and noise diagnostics. \\
    \bottomrule
    \end{tabularx}
\end{table*}

\section{Problem Formulation}\label{sec:problem}

\begin{figure}[!t]
    \centering
    \includegraphics[width=\columnwidth]{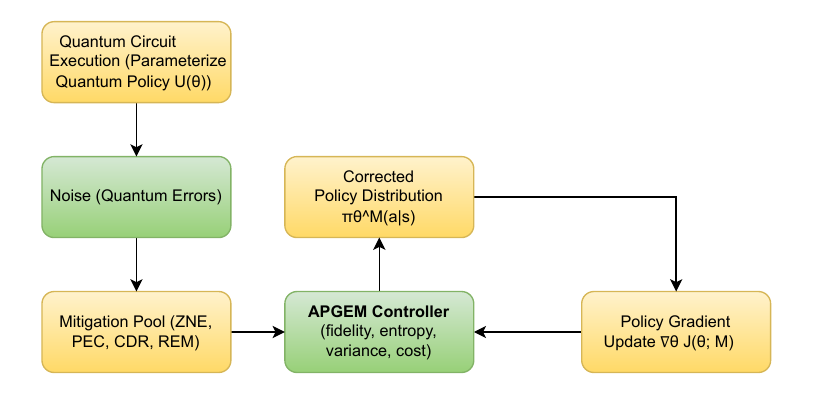}
    \caption{Block diagram of the Adaptive Policy-Guided Error Mitigation (APGEM) framework. Quantum circuits generate policies degraded by noise, which are corrected through an adaptive selection of mitigation strategies (ZNE, PEC, CDR, REM). The corrected policy guides QRL updates for solving the CVRP.}
    \label{fig:overview_apgem}
\end{figure}

This section introduces the \textbf{Capacitated Vehicle Routing Problem (CVRP)}, a fundamental extension of the Traveling Salesman Problem and a central challenge in combinatorial optimization. We present the mathematical formulation of the CVRP, followed by its reinforcement learning representation, and finally extend it to a Quantum Reinforcement Learning (QRL) framework. Special emphasis is placed on handling noisy quantum environments, where we integrate noise models and adaptive error mitigation strategies to preserve solution quality. An overview of the proposed APGEM framework for solving the CVRP under noisy quantum execution is shown in Fig.~\ref{fig:overview_apgem}.

\subsection{Capacitated Vehicle Routing Problem (CVRP)}

The Capacitated Vehicle Routing Problem (CVRP) can be formally defined on a weighted graph \(G=(V,E)\), where the set of nodes is given by
\[
V = \{0,1,2,\dots,n\}.
\]
The special node \(0\) denotes the depot, and the remaining nodes
\[
C := V \setminus \{0\}
\]
represent the set of \(n\) customers that require service. A fleet of identical vehicles is available, denoted by
\[
K = \{1,2,\dots,m\},
\]
where each vehicle has a homogeneous capacity \(Q > 0\). Each node \(i \in V\) is associated with a location vector \(x_i \in \mathbb{R}^2\), which encodes the geographical position of the depot and customers. The travel cost between any two nodes \(i, j \in V\) is expressed as
\[
c_{ij} = \lVert x_i - x_j \rVert_2
\]
in the Euclidean case, although more general cost metrics such as road distance or travel time may be used. The complete cost information is captured in the cost matrix
\[
C = [c_{ij}]_{i,j \in V}.
\]
Each customer \(i \in C\) is associated with a positive demand \(q_i > 0\), while the depot has no demand, i.e., \(q_0 := 0\). In addition, standard assumptions are often imposed: the cost matrix satisfies the triangle inequality, such that
\[
c_{ik} \leq c_{ij} + c_{jk}, \quad \forall i,j,k \in V,
\]
and self-loops are excluded by enforcing
\[
c_{ii} = +\infty, \quad \forall i \in V.
\]

A feasible solution to the CVRP, referred to as a \textit{routing plan}, consists of at most \(m\) vehicle tours, each corresponding to one vehicle in \(K\). Each tour must start and end at the depot, visit a subset of customers, and satisfy the requirement that every customer \(i \in C\) is visited exactly once by some vehicle. Furthermore, the total demand served on each vehicle route must respect the vehicle’s capacity constraint, such that for every vehicle \(k \in K\) we have
\begin{equation}
\label{eq:capacity}
\sum_{i \in \text{route}_k} q_i \leq Q.
\end{equation}
The objective of the CVRP is to minimize the total travel cost incurred across all vehicle tours, which can be mathematically formulated as
\begin{equation}
\label{eq:objective}
\min \sum_{k \in K} \sum_{(i,j) \in \text{route}_k} c_{ij}.
\end{equation}
This optimization problem ensures that the overall distance traveled by the fleet is minimized while simultaneously guaranteeing that all customer demands are satisfied and no vehicle exceeds its capacity.

Although the CVRP can be expressed as a mixed-integer programming (MIP) problem, solving it exactly is computationally intractable for large \(n\), since the number of feasible routing plans grows exponentially with the number of customers. For this reason, the classical formulation serves as the \textit{reference benchmark} for evaluating approximate approaches such as heuristics, metaheuristics, reinforcement learning (RL), and quantum reinforcement learning (QRL). In our framework, we do not directly solve the MIP; instead, the RL/QRL agents are designed to learn policies whose reward structure approximates the same cost minimization objective, thereby aligning the learning process with the classical CVRP formulation.

\subsection{Reinforcement Learning Formulation of the CVRP}

While the classical CVRP formulation provides the foundational optimization problem, in our framework the problem is cast into a reinforcement learning (RL) setting. This enables the use of learning-based agents, including quantum-enhanced reinforcement learning (QRL), to discover near-optimal routing policies without solving the intractable mixed-integer program directly. We now formalize the CVRP in the language of Markov Decision Processes (MDPs).

The RL environment for the CVRP is defined by the tuple \((\mathcal{S}, \mathcal{A}, \mathcal{P}, R, \gamma)\), where \(\mathcal{S}\) denotes the state space, \(\mathcal{A}\) the action space, \(\mathcal{P}\) the transition dynamics, \(R\) the reward function, and \(\gamma \in [0,1]\) the discount factor. In our setting, \(\gamma=1\), since all routes must terminate at the depot and the cost structure is episodic.

At each decision step, the environment provides a state vector
\[
s_t \in \mathcal{S},
\]
which encodes both vehicle-level and customer-level information. Specifically, the state is composed of:
\begin{itemize}
    \item \textbf{Vehicle features:} for each vehicle \(k \in K\), we maintain its current location \(l_k \in V\), its current load \(u_k \in [0,Q]\), and its normalized route length.
    \item \textbf{Customer features:} for each customer \(i \in C\), we encode its visited/unvisited status (binary indicator), its demand fraction \(q_i/Q\), and its spatial coordinates.
\end{itemize}
This representation ensures that the agent has full visibility of the routing progress, customer coverage, and residual capacities at every decision point.

At each step, the agent selects an action
\[
a_t \in \mathcal{A},
\]
corresponding to the choice of the next customer to visit. The action set is defined as
\begin{equation}
\label{eq:feasible}
\begin{split}
\mathcal{A}(s_t) = \{ i \in C :{}& \text{customer $i$ is unvisited}\\
&\text{and } u_k + q_i \leq Q \} \cup \{0\},
\end{split}
\end{equation}
where \(0\) represents the action of returning to the depot. The capacity feasibility check ensures that no action violates the vehicle capacity constraint. Once a customer is visited, it is marked as served and removed from the feasible action set for subsequent steps.

The environment updates deterministically according to the chosen action. If customer \(i \in C\) is selected, the active vehicle moves from its current location \(l_k\) to node \(i\), the load is updated as \(u_k \leftarrow u_k + q_i\), and the customer is marked as visited. If the depot action is chosen, the vehicle returns to node \(0\), the load is reset to \(0\), and a new vehicle may be dispatched until all customers are served or the vehicle limit \(m\) is reached.

The reward function is designed to approximate the classical CVRP objective of minimizing total routing cost. At each step, the agent receives an incremental reward
\begin{equation}
\label{eq:stepreward}
r_t = -c_{ij} + \beta \cdot \mathbf{1}_{\{\text{valid visit}\}} - \lambda \cdot \mathbf{1}_{\{\text{capacity violation}\}},
\end{equation}
where \(c_{ij}\) is the travel cost from the current location \(i\) to the chosen node \(j\), \(\beta > 0\) is a bonus for successfully visiting a customer, and \(\lambda \gg 0\) is a large penalty for infeasible actions such as attempting to exceed the vehicle capacity. At the end of an episode, an additional terminal penalty proportional to the number of unvisited customers is applied. The total return for an episode is therefore
\[
R = \sum_{t=0}^{T} r_t,
\]
which the agent seeks to maximize. Since step rewards are negative distances, maximizing return is equivalent to minimizing the total routing cost.

The objective of the RL (or QRL) agent is to learn a policy
\[
\pi_\theta : \mathcal{S} \to \Delta(\mathcal{A}),
\]
parameterized by \(\theta\), which maps states to probability distributions over feasible actions. The optimal policy \(\pi^*\) is defined as
\begin{equation}
\label{eq:optpolicy}
\pi^* = \arg \max_{\pi} \; \mathbb{E}_\pi \left[ \sum_{t=0}^{T} r_t \right],
\end{equation}
where the expectation is taken over the stochastic policy decisions and the environment dynamics. This formulation aligns directly with the classical CVRP objective, since maximizing cumulative reward corresponds to minimizing the total cost of serving all customers with feasible routes.

\subsection{Quantum Reinforcement Learning (QRL) Formulation of the CVRP}

Building on the reinforcement learning formulation of the CVRP, we now extend the agent to a quantum reinforcement learning (QRL) paradigm, where the policy network is parameterized by a quantum circuit. This approach leverages the representational capacity of quantum states to encode complex decision boundaries, while maintaining alignment with the same optimization objective as the classical CVRP.

In the QRL setting, the policy \(\pi_\theta\) is realized using a parameterized quantum circuit (PQC), often referred to as a variational quantum circuit. Let \(U(\theta)\) denote such a circuit with trainable parameters
\[
\theta = (\theta_1, \theta_2, \dots, \theta_p).
\]
The quantum state produced by the circuit is
\[
|\psi(\theta)\rangle = U(\theta) |0\rangle^{\otimes n_q},
\]
where \(n_q\) is the number of qubits. Measurement of this state in the computational basis yields a bitstring \(z \in \{0,1\}^{n_q}\) with probability
\[
P_\theta(z) = |\langle z | \psi(\theta) \rangle|^2.
\]

To interface with the CVRP environment, the measured bitstring is mapped to an action in the RL action space \(\mathcal{A}(s_t)\). This mapping is defined by a classical post-processing function
\[
a_t = f(z, s_t),
\]
which ensures that only feasible actions are selected, i.e.,
\[
a_t \in \mathcal{A}(s_t).
\]
In practice, the probability distribution over actions induced by the quantum circuit can be interpreted as
\begin{equation}
\label{eq:qpolicy}
\pi_\theta(a_t \mid s_t) = \sum_{z \in \mathcal{Z}(a_t)} P_\theta(z),
\end{equation}
where \(\mathcal{Z}(a_t)\) denotes the set of bitstrings mapped to action \(a_t\). This provides a quantum-native realization of the policy distribution.

The goal of QRL is to optimize the circuit parameters \(\theta\) such that the expected cumulative reward is maximized. Formally, the optimization objective is
\begin{equation}
\label{eq:objJ}
J(\theta) = \mathbb{E}_{\pi_\theta}\left[\sum_{t=0}^T r_t \right],
\end{equation}
where the expectation is taken over both the quantum measurement outcomes and the environment dynamics. Gradient-based updates are performed using parameter-shift rules, which allow unbiased estimation of gradients of quantum circuits with respect to their parameters. For a single parameter \(\theta_i\), the gradient of the expectation value of an observable \(O\) is given by
\[
\frac{\partial}{\partial \theta_i} \langle O \rangle_\theta = \frac{1}{2}\Big( \langle O \rangle_{\theta_i + \frac{\pi}{2}} - \langle O \rangle_{\theta_i - \frac{\pi}{2}} \Big).
\]
These gradient estimates are then used in stochastic gradient ascent to update the circuit parameters:
\[
\theta \leftarrow \theta + \eta \nabla_\theta J(\theta),
\]
where \(\eta > 0\) is the learning rate.

The overall objective of the QRL agent is to learn a quantum-parameterized policy \(\pi_\theta\) such that
\[
\pi^*_\theta = \arg \max_\theta \; \mathbb{E}_{\pi_\theta} \left[ \sum_{t=0}^T r_t \right].
\]
Since the reward function \(r_t\) directly encodes negative travel costs, maximizing this expected return is equivalent to minimizing the total routing cost, consistent with the classical CVRP objective:
\begin{equation}
\tag{\ref{eq:objective}}
\min \sum_{k \in K} \sum_{(i,j) \in \text{route}_k} c_{ij}.
\end{equation}

\subsection{Noise Model and Error Mitigation Techniques}\label{sec:noise-em}

Parameterized circuits are executed under explicit noise channels (depolarizing, dephasing, amplitude damping and readout noise), and four error mitigation techniques are considered: Zero-Noise Extrapolation (ZNE), Probabilistic Error Cancellation (PEC), Clifford Data Regression (CDR) and Readout Error Mitigation (REM). Each of these mitigation strategies addresses a distinct source of error: ZNE \cite{giurgica2020digital} reduces bias from gate noise, PEC \cite{mari2021extending} provides an unbiased but high-variance correction, CDR \cite{czarnik2021error} leverages classical data to improve non-Clifford circuit estimates, and REM \cite{nation2021scalable} specifically combats classical measurement noise. In practice, their performance depends on the noise profile of the underlying hardware and the tradeoff between statistical variance and systematic bias. In our framework, these methods form the foundation upon which the adaptive policy-guided error mitigation (APGEM) strategy is built, enabling dynamic selection of the most effective technique during QRL training.

\subsection{Adaptive Policy-Guided Error Mitigation (APGEM)}

While existing error mitigation techniques such as Zero-Noise Extrapolation (ZNE), Probabilistic Error Cancellation (PEC), Clifford Data Regression (CDR), and Readout Error Mitigation (REM) can improve quantum execution quality, their effectiveness depends strongly on the underlying noise characteristics, circuit structure, and execution conditions. A mitigation strategy that performs well under one noise regime may become suboptimal under another. Experimental observations further indicate that different techniques dominate under different operating conditions: no mitigation is often sufficient under very low noise, PEC and REM perform well under low-noise and readout-dominated settings, ZNE becomes advantageous under moderate noise, while CDR can be beneficial under severe noise conditions. Consequently, selecting a single mitigation strategy throughout training can lead to suboptimal performance.

To address this challenge, we propose the \textbf{Adaptive Policy-Guided Error Mitigation (APGEM)} framework, which formulates mitigation selection as an online contextual decision-making problem. Rather than statically applying one mitigation technique, APGEM continuously monitors the state of the learning process and dynamically selects the most suitable mitigation strategy according to the current execution context. Once a mitigation strategy $\mathcal{M}$ is selected, the corrected measurement statistics define the mitigated policy $\pi_\theta^{\mathcal{M}}(a_t|s_t)$, which is employed for policy-gradient updates,
\begin{equation}
\label{eq:pgmit}
\nabla_\theta J(\theta;\mathcal{M})=\mathbb{E}\left[\sum_{t=0}^{T}\nabla_\theta\log\pi_\theta^{\mathcal{M}}(a_t|s_t)\,G_t\right].
\end{equation}
The context representation, the cost-aware mitigation utility, the LinUCB contextual-bandit controller and the oracle reference used for evaluation are specified in Section~\ref{sec:apgem-method}.

\section{Methodology}\label{sec:methodology}
We propose a noise-resilient Quantum Reinforcement Learning (QRL) framework for the Capacitated Vehicle Routing Problem (CVRP), enhanced with Adaptive Policy-Guided Error Mitigation (APGEM). The framework integrates a hard-feasible CVRP environment, a variational quantum policy trained by a policy-gradient estimator, realistic noise modeling, and a context-aware mitigation controller into a single pipeline, illustrated in Fig.~\ref{fig:apgem-framework}. The components are described below, and every quantity stated here corresponds to the released implementation.

\subsection{Proposed Framework}
\begin{figure}[!t]
    \centering
    \includegraphics[width=\columnwidth]{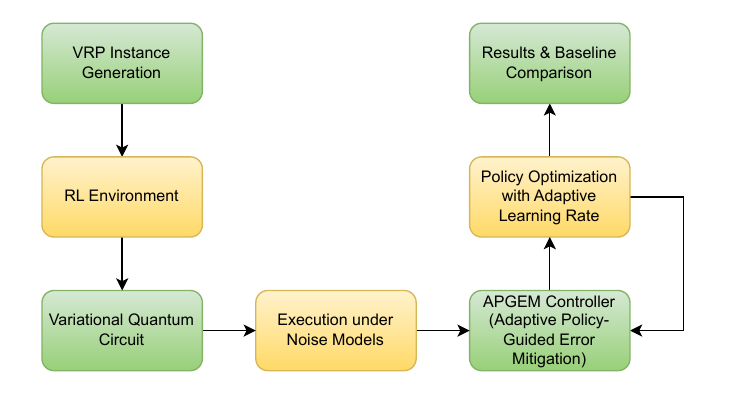}
    \caption{Block diagram of the proposed methodology. The workflow begins with VRP instance generation, followed by reinforcement learning environment interaction through a variational quantum circuit. Execution under noise is adaptively corrected by the APGEM controller, with fidelity and entropy metrics guiding policy optimization. The training loop produces results compared against a classical baseline.}
    \label{fig:apgem-framework}
\end{figure}

The proposed framework integrates four components into a single closed-loop pipeline (Fig.~\ref{fig:apgem-framework}): a hard-feasible CVRP environment, a variational quantum policy, a realistic noise model, and the APGEM mitigation controller. At each step the quantum policy proposes an action under noisy execution; the APGEM controller selects a mitigation technique from its pool based on the current operating context; the mitigated measurement distribution defines the action policy; and the resulting reward drives both the policy-gradient update of the quantum circuit and the online update of the mitigation controller. The CVRP environment, its state and action spaces, and the shaped reward are defined in the following subsection, and the remaining components are detailed thereafter.

\subsection{CVRP as a Reinforcement Learning Environment}
The CVRP is encoded as an episodic reinforcement learning problem. A state $s_t$ describes the partial solution at step $t$: the current node, the remaining vehicle capacity, the set of visited customers, and the number of vehicles used. The action space corresponds to selecting the next customer or returning to the depot, and capacity feasibility is enforced as a hard constraint by an action mask that removes any customer whose demand would exceed the residual capacity.

The reward is shaped to align cumulative return with minimal routing cost while rewarding progress and discouraging infeasibility:
\begin{equation}
\label{eq:shapedreward}
r_t =
\begin{cases}
-c_{ij} + \beta, & \text{on serving a feasible customer}\\ & \text{$j$ from $i$},\\[2pt]
-c_{i,\mathrm{depot}}, & \text{on returning to the depot from } i,\\[2pt]
-\lambda, & \text{on an infeasible attempt}\\ & \text{(masked out in practice)},
\end{cases}
\end{equation}
with a terminal penalty $-\lambda\,|\mathcal{U}|$ proportional to the number of customers $\mathcal{U}$ left unserved at episode end. We use $\beta=5$ and $\lambda=100$. The visit bonus $\beta$ provides a dense progress signal, and the distance terms recover the classical objective, so that maximizing return corresponds to minimizing total distance over a feasible solution.

\paragraph{Feasibility-guaranteed decoding.}
To ensure that every episode yields a complete, capacity-feasible solution, the environment uses a coverage-preserving construction: it never terminates while customers remain unserved and opens a new route whenever no active route has capacity for the chosen customer, with a vehicle budget large enough to admit a worst-case one-route-per-customer solution. As a safety net, a cheapest-insertion repair inserts any residual unserved customer at its lowest-cost capacity-feasible position, opening a single-customer route only if required. Consequently the reported solutions serve every customer exactly once and respect capacity by construction, and the reported feasibility rate reflects this guarantee rather than a quantity the policy fails to reach.

\subsection{Quantum Policy Representation}
The routing policy is parameterized by a variational quantum circuit (VQC) $U(\theta)$. For an instance with $n$ customers we use $n_q=\lceil \log_2(n+1)\rceil$ qubits, capped for NISQ scale. The circuit applies a single-qubit $R_y$ feature-encoding layer that loads a compact state description (fraction visited, load fraction, normalized nearest-customer distance, vehicle fraction), followed by $L$ variational layers, each a linear controlled-$Z$ entangling pattern followed by trainable $R_y$ rotations with parameters $\theta \in \mathbb{R}^{n_q L}$. Measurement in the computational basis yields a distribution over bitstrings $z \in \{0,1\}^{n_q}$,
\begin{equation}
\label{eq:pz}
p_z(\theta \mid s) =
\mathrm{Tr}\!\left( \Pi_z \, U(\theta,s) \lvert 0 \rangle \langle 0 \rvert^{\otimes n_q} U^\dagger(\theta,s) \right).
\end{equation}

\paragraph{From measurements to feasible actions.}
At each step the feasible actions are placed in a canonical, parameter-independent order (customers by ascending distance from the current node, depot last), and the $2^{n_q}$ outcome distribution is aggregated into $k$ contiguous blocks, one per feasible action. This yields a categorical policy over the feasible action set,
\begin{equation}
\label{eq:policybin}
\pi_\theta(a \mid s) = \sum_{z \in \mathcal{B}_a} p_z(\theta \mid s),
\end{equation}
where $\mathcal{B}_a$ is the contiguous block assigned to action $a$. The distance-ordered, contiguous assignment gives the circuit a consistent and learnable correspondence between measurement mass and routing decisions, so that concentrating probability on low-index outcomes corresponds to selecting nearer customers. Actions are sampled from $\pi_\theta$ during training and taken greedily at evaluation.

\subsection{Policy Optimization}
The policy parameters are trained with the REINFORCE policy-gradient estimator and exact parameter-shift gradients. For a sampled trajectory with rewards $\{r_t\}$ and return-to-go $G_t=\sum_{t'\ge t}\gamma^{\,t'-t} r_{t'}$ ($\gamma=0.99$), the gradient of the expected return is
\begin{equation}
\label{eq:reinforce}
\nabla_\theta J
= \mathbb{E}\!\left[\sum_t (G_t-\bar{G})\,\nabla_\theta \log \pi_\theta(a_t\mid s_t)\right],
\end{equation}
with a mean baseline $\bar{G}$ and returns standardized across a small batch of trajectories for variance reduction. Because $\pi_\theta(a\mid s)$ is a sum of circuit output probabilities, its gradient is obtained exactly through the parameter-shift rule,
\begin{equation}
\label{eq:psrpz}
\frac{\partial p_z}{\partial \theta_j}
= \tfrac{1}{2}\!\left(p_z(\theta_j+\tfrac{\pi}{2}) - p_z(\theta_j-\tfrac{\pi}{2})\right),
\end{equation}
which holds for the $R_y$-generated parameters of the ansatz. Parameters are updated by ascent $\theta \leftarrow \theta + \eta\,\nabla_\theta J \bmod 2\pi$ with learning rate $\eta$. This estimator is unbiased for the expected return and provides a directed learning signal, in contrast to perturbation heuristics whose expected update is zero.

\subsection{Noise Modeling}
To emulate near-term hardware, the policy circuit is executed under a completely positive trace-preserving map $\mathcal{N}$, $\rho(\theta)=\mathcal{N}\!\big(U(\theta)\lvert 0\rangle\langle 0\rvert^{\otimes n_q}U^\dagger(\theta)\big)$. We implement the noise families of Section~\ref{sec:setup} (depolarizing, dephasing, amplitude damping, two-qubit gate, and readout); in particular, depolarizing noise $\rho \mapsto (1-p)\rho + p\,I/2^{n_q}$; dephasing $\rho \mapsto (1-p)\rho + p\,Z\rho Z$; amplitude damping via Kraus operators $K_0,K_1$; and readout error described by a confusion matrix acting on the measured probabilities. In the adaptive-mitigation experiments the depolarizing level $p$ and the readout error rate $r$ are varied independently across operating regimes so that the mitigation controller is exposed to changing conditions.

\subsection{Error Mitigation Pool}
The candidate mitigation pool is $\mathcal{M}=\{\text{NoMitigation},\allowbreak\ \text{ZNE},\allowbreak\ \text{PEC},\allowbreak\ \text{CDR},\allowbreak\ \text{REM}\}$. Zero-Noise Extrapolation evaluates the observable at scaled noise levels $\{\lambda_i\}$ and extrapolates to the zero-noise limit, $\hat{E}(0)=\sum_i c_i E(\lambda_i)$. Probabilistic Error Cancellation writes each ideal gate as a quasi-probability mixture $\mathcal{G}=\sum_j \alpha_j\tilde{\mathcal{G}}_j$ and reweights sampled noisy gates, at the cost of increased variance. Clifford Data Regression fits a correction $y\approx ax+b$ from near-Clifford proxy circuits that are classically simulable. Readout Error Mitigation inverts a calibrated confusion matrix, $\hat{p}=A^{-1}\tilde{p}$, with regularization for large systems. Each technique carries a distinct sampling cost, which we quantify in shots and use in the mitigation utility below.

\subsection{Adaptive Policy-Guided Error Mitigation}\label{sec:apgem-method}
No single mitigation technique is optimal across all conditions: unmitigated execution suffices at very low noise, PEC and ZNE lead at low-to-moderate depolarizing noise, readout mitigation is preferable under readout-dominated noise, and CDR is preferable at severe noise. APGEM therefore formulates mitigation selection as an online contextual decision.

\paragraph{Context representation.}
At step $t$ APGEM forms a base context
\begin{equation}
\label{eq:context}
x_t^{\mathrm{base}}=(p_t, r_t, F_t, \Delta F_t, H_t, d_t, B_t, AR_t, RT_t),
\end{equation}
comprising the depolarizing level, readout rate, unmitigated fidelity, fidelity degradation $\Delta F_t=1-F_t$, policy entropy, normalized circuit depth, remaining shot budget, approximation ratio, and recent reward trend, each normalized to $[0,1]$ (noise and readout by their maximum sampled values). Because the optimal technique is a non-monotonic function of the noise level (for example, ZNE is best in the middle of the range rather than at the extremes), a purely linear controller in $x_t^{\mathrm{base}}$ cannot separate the regimes. We therefore augment the context with radial-basis features centred on each regime,
\begin{equation}
\label{eq:rbf}
\phi_c(u)=\exp\!\big(-((u-c)/w)^2\big),
\end{equation}
using centres $c\in\{0.05,0.15,0.25,0.40,0.60,0.90\}$ with width $w=0.12$ on the normalized depolarizing level and $c\in\{0.0,0.5\}$ with width $w=0.18$ on the normalized readout rate. The full context $x_t\in\mathbb{R}^{17}$ concatenates the nine base features with these eight basis features, making each technique's optimal region linearly separable.

\paragraph{Mitigation utility.}
For each strategy $\mathcal{M}$ the utility trades fidelity against sampling cost and estimator variance,
\begin{equation}
\label{eq:utility}
U(\mathcal{M};x_t)=\sigma\!\big(k(F_{\mathcal{M}}-F_0)\big)-\alpha\,C_{\mathcal{M}}-\beta\,V_{\mathcal{M}},
\end{equation}
where $F_{\mathcal{M}}$ is the fidelity after applying $\mathcal{M}$, $\sigma$ is the logistic function, and $F_0$ is a fixed usability threshold ($F_0=0.80$, $k=12$). Centring the sigmoid at a usability threshold rather than at the unmitigated fidelity makes a fidelity gain near the threshold far more valuable than a gain in the already-saturated high-fidelity region, so an expensive technique is rewarded only when it lifts fidelity across the usable regime. The cost $C_{\mathcal{M}}$ is the per-application shot count normalized to the most expensive technique, giving $C_{\text{NoMitigation}}\!=\!0.06$, $C_{\text{ZNE}}\!=\!0.12$, $C_{\text{REM}}\!=\!0.25$, $C_{\text{PEC}}\!=\!0.31$ and $C_{\text{CDR}}\!=\!1.0$; $V_{\mathcal{M}}$ is the estimator variance; and $\alpha=0.15$, $\beta=0.10$ weight the cost-performance trade-off.

\paragraph{Contextual-bandit controller.}
APGEM learns mitigation preferences with a LinUCB contextual bandit. For each arm $\mathcal{M}$ it maintains $A_{\mathcal{M}}$ and $b_{\mathcal{M}}$, initialized as $A_{\mathcal{M}}=I_d$ (ridge regularization) and $b_{\mathcal{M}}$ with a small optimistic offset to encourage early sampling, and estimates $\hat{\theta}_{\mathcal{M}}=A_{\mathcal{M}}^{-1}b_{\mathcal{M}}$. Given $x_t$, the score is
\begin{equation}
\label{eq:linucb}
s_{\mathcal{M}}=\hat{\theta}_{\mathcal{M}}^{\top}x_t + \lambda\sqrt{x_t^{\top}A_{\mathcal{M}}^{-1}x_t},
\end{equation}
with a fixed, non-decaying exploration coefficient $\lambda$ so that exploration persists wherever the per-arm posterior remains uncertain. The selected strategy is $\mathcal{M}_t=\arg\max_{\mathcal{M}} s_{\mathcal{M}}$. After application, the observed reward
\begin{equation}
\label{eq:banditreward}
R_t=\sigma\!\big(k(F_{\mathcal{M}_t}-F_0)\big)-\alpha\,C_{\mathcal{M}_t}-\beta\,V_{\mathcal{M}_t}
\end{equation}
is used to update the chosen arm only, via the Sherman-Morrison rank-one update of $A_{\mathcal{M}_t}^{-1}$, $A_{\mathcal{M}_t}\!\leftarrow\!A_{\mathcal{M}_t}+x_tx_t^{\top}$ and $b_{\mathcal{M}_t}\!\leftarrow\!b_{\mathcal{M}_t}+R_t x_t$. Only the chosen arm's reward enters the update, so no counterfactual or oracle information leaks into learning. At deployment the controller acts greedily, $\arg\max_{\mathcal{M}}\hat{\theta}_{\mathcal{M}}^{\top}x_t$. The mitigated distribution of the selected technique defines the policy $\pi_\theta^{\mathcal{M}_t}(a_t\mid s_t)$ used for action sampling and policy optimization, so mitigation selection is an adaptive component of the learning loop rather than a static preprocessing step. The complete procedure is given in Algorithm~\ref{alg:apgem_linucb}.

\begin{algorithm}[!t]
\caption{Adaptive Policy-Guided Error Mitigation (APGEM) using LinUCB}
\label{alg:apgem_linucb}
\begin{algorithmic}[1]
\State \textbf{Input:} pool $\mathcal{M}=\{\text{NoMitigation},\text{ZNE},\text{PEC},\text{CDR},\text{REM}\}$; exploration coefficient $\lambda$; usability threshold $F_0$; weights $\alpha,\beta$
\ForAll{$m \in \mathcal{M}$}
    \State $A_m \leftarrow I_d$; \quad $b_m \leftarrow b_0$ \Comment{optimistic init}
\EndFor
\For{each training step $t$}
    \State Form context $x_t$ (base features $+$ RBF basis)
    \ForAll{$m \in \mathcal{M}$}
        \State $\hat{\theta}_m \leftarrow A_m^{-1}b_m$
        \State $s_m \leftarrow \hat{\theta}_m^{\top} x_t + \lambda\sqrt{x_t^{\top} A_m^{-1} x_t}$
    \EndFor
    \State $m_t \leftarrow \arg\max_{m} s_m$; \quad apply $m_t$; \quad observe $F_{m_t}, C_{m_t}, V_{m_t}$
    \State $R_t \leftarrow \sigma(k(F_{m_t}-F_0)) - \alpha C_{m_t} - \beta V_{m_t}$
    \State $A_{m_t} \leftarrow A_{m_t} + x_t x_t^{\top}$; \quad $b_{m_t} \leftarrow b_{m_t} + R_t x_t$
\EndFor
\end{algorithmic}
\end{algorithm}

\subsection{Oracle Reference and Regret}
As an upper bound we define a per-context oracle that selects the strategy with the highest expected utility,
\begin{equation}
\label{eq:oracle}
m^\ast(x_t)=\arg\max_{\mathcal{M}}\ \mathbb{E}\!\left[U(\mathcal{M};x_t)\right],
\end{equation}
where the expectation is estimated by averaging over repeated executions at the same context. The expected-utility oracle is the correct reference for regret, because an argmax taken over a single stochastic realization overstates the achievable performance by an irreducible amount and would yield non-vanishing regret even for an optimal policy. We therefore report APGEM's regret against the expected-utility oracle and separately report the residual stochastic gap, so that policy suboptimality is distinguished from irreducible sampling variance. The oracle computation is summarized in Algorithm~\ref{alg:oracle}.

\begin{algorithm}[!t]
\caption{Oracle Mitigation Policy (Expected Utility)}
\label{alg:oracle}
\begin{algorithmic}[1]
\For{each context $x_t$}
    \ForAll{$m \in \mathcal{M}$}
        \State Estimate $\bar{U}(m;x_t) \leftarrow \tfrac{1}{N}\sum_{i=1}^{N} U^{(i)}(m;x_t)$
    \EndFor
    \State $m^\ast \leftarrow \arg\max_m \bar{U}(m;x_t)$
\EndFor
\end{algorithmic}
\end{algorithm}

By integrating feasibility-guaranteed decoding, a policy-gradient-trained quantum policy, explicit noise modeling, and a context-aware mitigation controller, the framework turns error mitigation into an adaptive learning component that selects context-dependent strategies across diverse NISQ operating regimes while optimizing a feasible routing policy.

\section{Performance Metrics}
\label{sec:metrics}

The proposed framework integrates Quantum Reinforcement Learning (QRL), variational quantum circuits (VQCs), and Adaptive Policy-Guided Error Mitigation (APGEM). To evaluate both quantum reliability and optimization effectiveness, we employ four categories of metrics: (i) quantum execution quality, (ii) optimization and learning performance, (iii) adaptive mitigation effectiveness, and (iv) robustness and statistical reliability. All metrics are computed from the measured output probability distributions of the executed circuits, since the framework operates on measurement statistics rather than reconstructed density matrices.

The metrics and their definitions are summarized in Table~\ref{tab:metrics}.

\begin{table*}[!t]
    \centering
    \caption{Summary of performance metrics used in evaluating APGEM.}
    \label{tab:metrics}
    \renewcommand{\arraystretch}{1.1}
    \footnotesize
    \begin{tabularx}{\textwidth}{@{}p{3.6cm}p{5.6cm}X@{}}
    \toprule
    \textbf{Metric} & \textbf{Definition / Formula} & \textbf{Interpretation} \\
    \midrule

    \textbf{Distribution Fidelity ($F$)} &
    $F(p,q)=\big(\sum_i \sqrt{p_i q_i}\big)^2$ &
    Classical (Bhattacharyya) fidelity between mitigated and ideal output distributions; higher is better. \\

    \textbf{Normalized Entropy ($H$)} &
    $H(p)=-\tfrac{1}{n_q}\sum_i p_i\log_2 p_i$ &
    Shannon entropy of the measured distribution, normalized to $[0,1]$; lower indicates a sharper, less noise-dominated distribution. \\

    \textbf{Cumulative Reward ($R$)} &
    $R=\sum_t r_t$ &
    Accumulated shaped reward over an episode; higher corresponds to lower-cost feasible routing. \\

    \textbf{Routing Cost ($C$)} &
    Total travel distance of all routes (with depot returns) &
    Domain-specific solution quality on the feasibility-guaranteed decode; lower is better. \\

    \textbf{Approximation Ratio ($AR$)} &
    $AR_{\text{BKS}}=\dfrac{C_{\text{quantum}}}{C_{\text{BKS}}}$ (and vs.\ best classical) &
    Cost relative to best-known (primary) and to the best classical baseline; values near $1$ are near-optimal. \\

    \textbf{Mitigation Gain ($\Delta M$)} &
    $M_{\text{mitigated}}-M_{\text{unmitigated}}$ &
    Execution-quality improvement from mitigation at matched conditions. \\

    \textbf{Technique Selection Frequency} &
    Normalized distribution over $\{\text{NoMitigation},\text{ZNE},\text{PEC},\text{CDR},\text{REM}\}$ &
    Characterizes APGEM adaptivity and the preferred technique per regime. \\

    \textbf{Average Utility ($U$)} &
    $\sigma(k(F_{\mathcal M}-F_0))-\alpha C_{\mathcal M}-\beta V_{\mathcal M}$, $F_0{=}0.80$ &
    Cost-aware trade-off of fidelity gain, normalized shot cost, and variance used by APGEM. \\

    \textbf{Mitigation Overhead ($\Omega$)} &
    $\dfrac{\text{Shots}_{\mathcal M}}{\text{Shots}_{\text{NoMitigation}}}$ &
    Multiplicative sampling cost relative to unmitigated execution. \\

    \textbf{Oracle Gap} &
    $\dfrac{U_{\text{oracle}}-U_{\text{APGEM}}}{U_{\text{oracle}}}\times 100$ &
    Distance to the expected-utility oracle; lower is better. \\

    \textbf{Policy Regret} &
    $\sum_{t}[\bar U^{*}(x_t)-\bar U_{\text{APGEM}}(x_t)]$ &
    Expected-utility gap to the oracle over time, with the stochastic floor reported separately. \\

    \textbf{Performance Stability} &
    Mean $\pm$ bootstrap $95\%$ CI and standard deviation across seeds &
    Consistency under random seeds and noise realizations. \\

    \textbf{Statistical Significance} &
    Friedman, Nemenyi (CD), Wilcoxon, Cliff's delta &
    Whether differences between methods or strategies are statistically meaningful. \\

    \bottomrule
    \end{tabularx}
\end{table*}

\section{Experimental Setup}\label{sec:setup}
We evaluate the proposed framework on the Capacitated Vehicle Routing Problem (CVRP) using standard benchmark instances rather than a single synthetic case. The study uses instances from the CVRPLIB library, dominated by the Augerat Set~A (25 instances) together with representatives from the B, E, F, M and P families, spanning 15 to 100 customers. Each instance specifies a single depot, fixed customer demands, and a fixed vehicle capacity read directly from the instance file; inter-node distances are the Euclidean (EUC\_2D) distances of the benchmark, rounded to integers as in the CVRPLIB convention. Every instance is paired with its published best-known solution (BKS), which serves as the reference for the approximation ratio. This benchmark-based setup replaces single-instance evaluation and provides the blocking factor for the statistical tests reported in Section~\ref{sec:metrics}.

\subsection{Classical Baselines}
The implementation includes a suite of constructive and metaheuristic CVRP solvers: Nearest Neighbour, Savings (Clarke--Wright), Genetic Algorithm, Simulated Annealing, and Tabu Search, together with the Google OR-Tools routing solver as a strong reference. Each solver returns a set of vehicle routes and a total cost, and a unified wrapper runs all baselines and aggregates the results, reporting the median over random seeds for the stochastic methods. Capacity feasibility is enforced during route construction and perturbation, and the total cost is the sum of pairwise arc distances including depot returns.

\subsection{Quantum Policy and Circuit Design}
\begin{figure}[!t]
    \centering
    \includegraphics[width=\columnwidth]{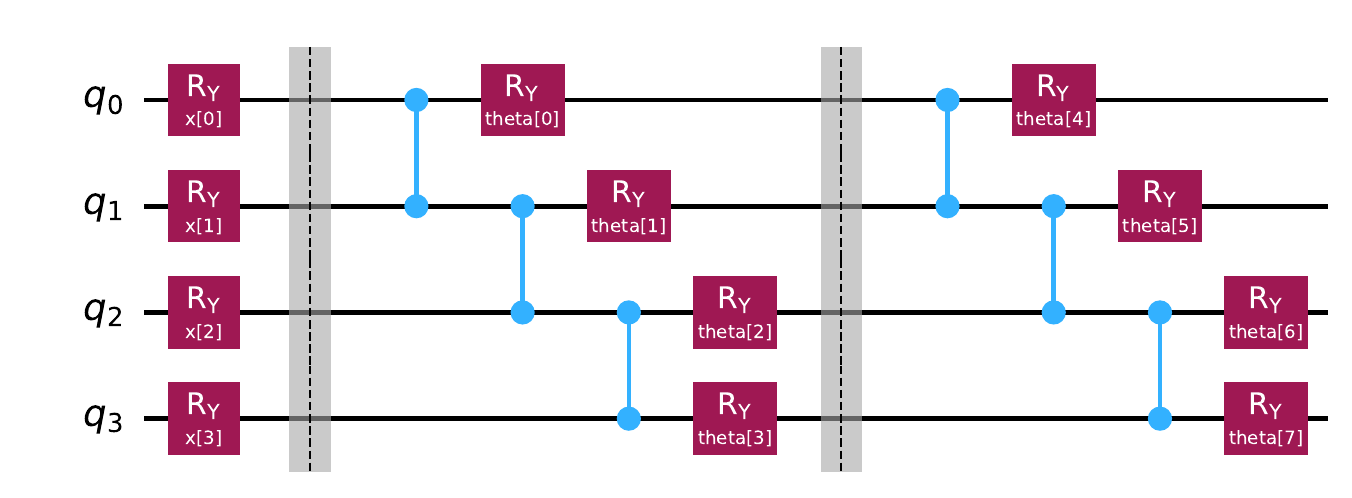}
    \caption{Quantum policy ansatz for the CVRP--QRL framework. The first layer applies $R_y(\theta_i)$ rotations to encode environment-state features into the quantum register. This is followed by a linear entanglement layer of controlled-$Z$ gates to capture correlations across qubits. Two variational layers are then stacked, each consisting of trainable $R_y(\phi^{(k)}_i)$ rotations and entangling operations, allowing expressive state transformations. The final measurement distribution guides action selection within the reinforcement learning loop.}
    \label{fig:circuit}
\end{figure}

The routing policy is a parameterized quantum circuit (PQC) evaluated within an episodic reinforcement learning loop over a custom CVRP environment. For an instance with $n$ customers the circuit uses $n_q=\lceil\log_2(n+1)\rceil$ qubits, capped for NISQ scale. The circuit has three components, illustrated in Fig.~\ref{fig:circuit}: (i) a feature-encoding layer of single-qubit $R_y(\theta_i)$ rotations that loads a compact state description into the register, (ii) a linear entanglement layer of controlled-$Z$ gates, and (iii) a variational block of trainable $R_y(\phi_i^{(k)})$ rotations repeated over $L$ layers, each layer preceded by the entangling pattern. This manually specified ansatz balances expressive power against circuit depth for noisy intermediate-scale quantum (NISQ) execution. Circuits are simulated on the Qiskit AerSimulator, and parameters are initialized at random.

\subsection{Reinforcement Learning Environment}
The environment encodes the CVRP as an episodic decision process. The state is a compact feature vector comprising the fraction of customers already served, the current vehicle load as a fraction of capacity, the normalized distance to the nearest unvisited customer, and the fraction of the vehicle budget used; these features are mapped to encoding angles in $[0,\pi]$ for the $R_y$ layer. The action set consists of the unvisited customers whose demand fits the residual capacity, plus a return-to-depot action; capacity feasibility is enforced as a hard constraint by an action mask. Step rewards combine the negative arc distance with a per-visit progress bonus, and a terminal penalty is applied for any unserved customers. To guarantee complete feasible solutions, the environment uses a coverage-preserving construction that opens a new route whenever no active route has capacity, followed by a cheapest-insertion repair; consequently every episode yields a solution that serves each customer exactly once within capacity. The measurement distribution of the (mitigated) circuit is aggregated over the feasible actions in a distance-ordered, contiguous mapping to form the categorical policy, and the policy parameters are trained by the REINFORCE policy-gradient estimator with exact parameter-shift gradients. During training the environment resets at the start of each episode, and per-episode metrics are recorded, including mean reward, distribution fidelity, normalized entropy, approximation ratio, total quantum route cost, and best classical route cost, with execution time logged for both the quantum policy and the classical solvers.

\subsection{Noise Modeling}
To replicate NISQ conditions, circuits are executed on Aer with explicit noise models across several severity levels ($0.01$, $0.05$, $0.08$, $0.10$). Five noise families are supported: depolarizing, amplitude damping, phase damping (dephasing), two-qubit gate noise (depolarizing on the entangling gates), and measurement\slash readout noise, each instantiated at the defined levels to yield a grid of scenarios. For the adaptive-mitigation experiments the depolarizing level is additionally swept over a wider range ($0.01$ to $0.18$) and the readout error rate is varied independently, so that the mitigation controller is exposed to distinct operating regimes rather than a single fixed noise level.

\subsection{Adaptive Error Mitigation}
\begin{figure*}[!t]
    \centering
    \includegraphics[width=\textwidth]{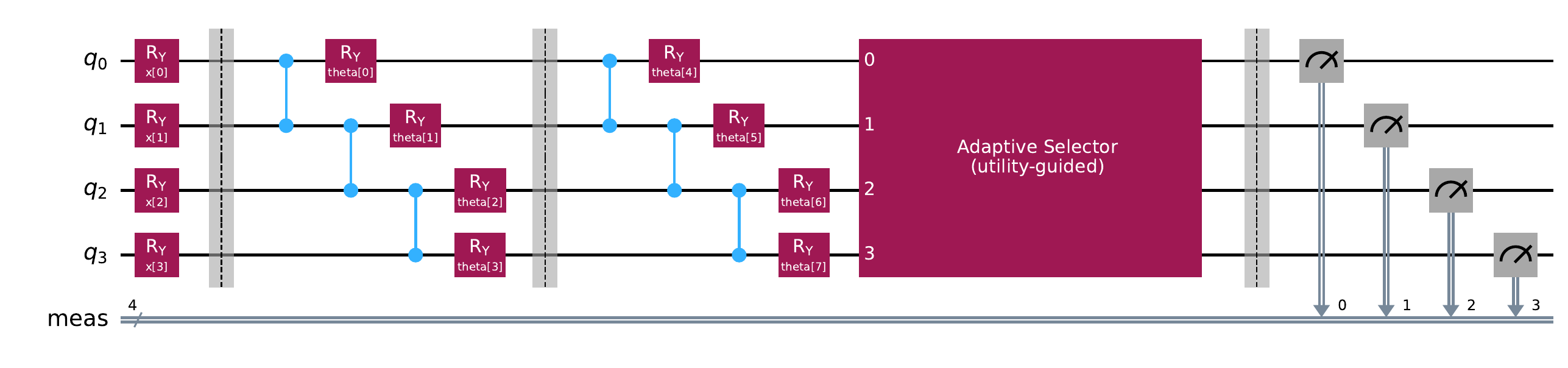}
    \caption{Circuit-level illustration of the adaptive error mitigation module integrated into the CVRP--QRL framework.}
    \label{fig:adaptive_circuit}
\end{figure*}

An Adaptive Policy-Guided Error Mitigation (APGEM) module selects among mitigation techniques at run time. The candidate pool is $\{\text{NoMitigation},\allowbreak\ \text{ZNE},\allowbreak\ \text{PEC},\allowbreak\ \text{CDR},\allowbreak\ \text{REM}\}$, and the selection is made by a LinUCB contextual-bandit controller from a context vector summarizing the current operating regime (depolarizing level, readout rate, unmitigated fidelity and its degradation, entropy, normalized circuit depth, remaining shot budget, approximation ratio, and recent reward trend), augmented with radial-basis features on the noise and readout axes. Each technique is scored by a cost-aware utility
\begin{equation}
\tag{\ref{eq:utility}}
U(\mathcal{M};x_t)=\sigma\!\big(k(F_{\mathcal{M}}-F_0)\big)-\alpha\,C_{\mathcal{M}}-\beta\,V_{\mathcal{M}},
\end{equation}
which rewards mitigated distribution fidelity $F_{\mathcal{M}}$ relative to a fixed usability threshold $F_0$ while penalizing the normalized sampling cost $C_{\mathcal{M}}$ and estimator variance $V_{\mathcal{M}}$. The module tracks per-technique usage counts, recent performance, and the observed utility, and updates the chosen arm online. Figure~\ref{fig:adaptive_circuit} shows a circuit-level illustration of the mitigation strategies: ZNE by stretching entangling gates, PEC through stochastic inverse gates, CDR via calibrated near-Clifford runs, and REM by correcting measurement statistics, with the adaptive selector choosing the technique at runtime according to the utility.

\subsection{Evaluation Protocol}
The evaluation reports: (i) total route cost; (ii) approximation ratio, computed primarily against the best-known solution and secondarily against the best classical baseline; (iii) feasibility, which is guaranteed by the coverage construction and additionally reported as a served-customer fraction; (iv) mean reward per episode; (v) distribution fidelity and normalized entropy as indicators of quantum-execution quality; and (vi) runtime for both the quantum policy and the classical solvers. Results are reported across solvers, episodes, noise conditions, and mitigation techniques, and differences are assessed with non-parametric significance tests (Friedman, Nemenyi post-hoc, and Wilcoxon signed-rank) using the instance as the blocking factor, with bootstrap confidence intervals for stability.

\section{Results and Discussion}\label{sec:results}
This section reports the empirical evaluation of the proposed framework. We first validate the classical baseline suite on a thirty-instance CVRPLIB benchmark with formal significance testing, then establish that the quantum routing policy is feasible and learns, and report its routing performance and scaling honestly. The core contribution, the Adaptive Policy-Guided Error Mitigation (APGEM) controller, is then evaluated in detail: we show that it selects different mitigation techniques under different operating conditions, approaches an expected-utility oracle, and significantly outperforms fixed strategies. All figures and tables are produced directly by the released code, and every reported number is taken from the generated result files.

\subsection{Experimental Configuration}
All experiments were executed on the software stack in Table~\ref{tab:env}, recorded automatically by the reproducibility logger. Mitigation techniques are implemented natively and the critical-difference computation uses a built-in studentized-range table, so no external mitigation or post-hoc libraries are required.

\begin{table}[!t]
    \centering
    \caption{Software environment used for all reported results.}
    \label{tab:env}
    \footnotesize
    \setlength{\tabcolsep}{3pt}
    \begin{tabular}{ll}
    \toprule
    Component & Version \\
    \midrule
    Python        & 3.10.11 \\
    NumPy         & 2.2.6 \\
    SciPy         & 1.14.1 \\
    pandas        & 2.2.3 \\
    Qiskit        & 1.4.0 \\
    Qiskit Aer    & 0.16.1 \\
    OR-Tools      & 9.15.6755 \\
    scikit-learn  & 1.7.1 \\
    \bottomrule
    \end{tabular}
\end{table}

The benchmark comprises thirty CVRPLIB instances (twenty-five from the Augerat Set~A, plus one representative each from the B, E, F, M and P families), spanning 15 to 100 customers, each paired with its published best-known solution (BKS).

\subsection{Classical Baseline Benchmark and Statistical Validation}
\begin{figure}[!t]
    \centering
    \includegraphics[width=\columnwidth]{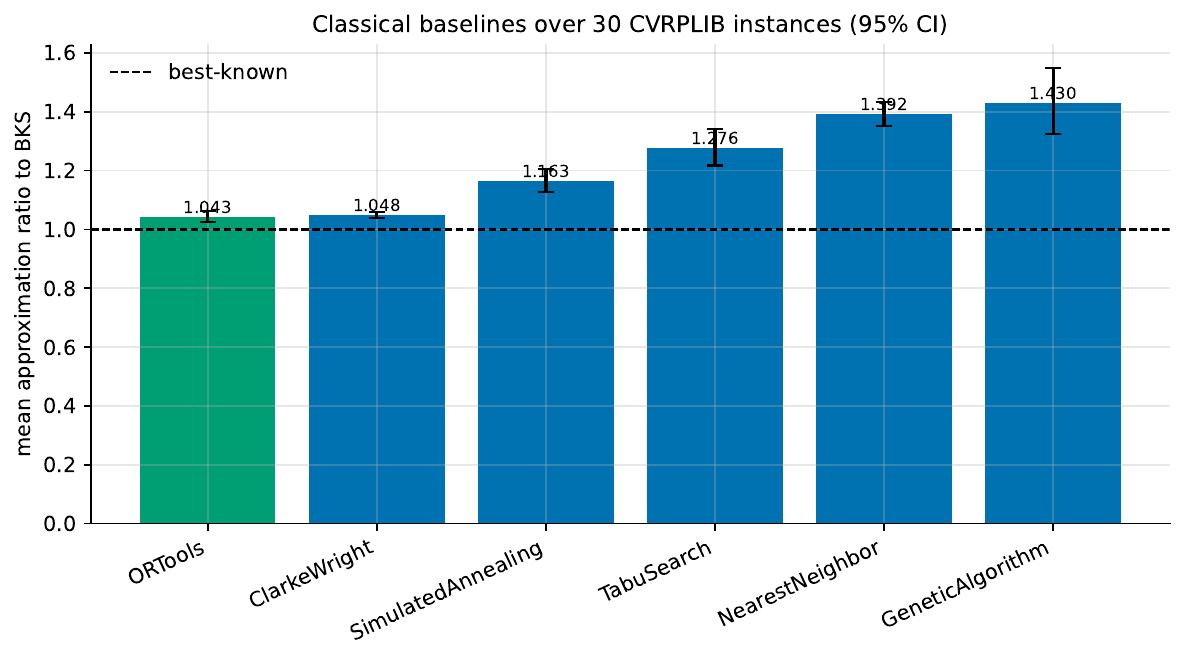}
    \caption{Mean approximation ratio to best-known with bootstrap $95\%$ confidence intervals over 30 CVRPLIB instances. OR-Tools and Clarke--Wright are the strongest and statistically indistinguishable.}
    \label{fig:benchmark_ci}
\end{figure}

\begin{figure}[!t]
    \centering
    \includegraphics[width=\columnwidth]{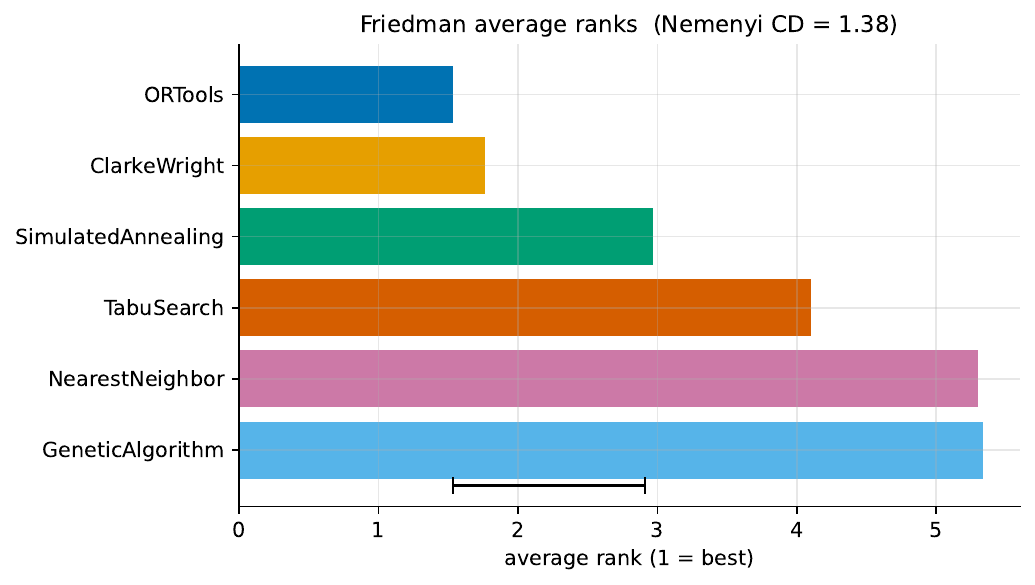}
    \caption{Friedman average ranks (1 = best) with the Nemenyi critical difference ($\text{CD}=1.38$). OR-Tools and Clarke--Wright group together and dominate the metaheuristics.}
    \label{fig:benchmark_cd}
\end{figure}

Table~\ref{tab:benchmark} reports the approximation ratio to best-known for each classical solver over the thirty instances, with the instance used as the blocking factor. A Friedman omnibus test rejects the null of equal performance with $\chi^2 = 121.47$ and $p = 1.53\times10^{-24}$, and the Nemenyi post-hoc critical difference is $1.38$ average-rank units. OR-Tools attains the best mean approximation ratio ($1.043$) and rank ($1.53$), closely followed by Clarke--Wright ($1.048$, rank $1.77$); the two are not statistically distinguishable (Wilcoxon $p = 0.50$) and fall within the critical difference, while both significantly outperform the metaheuristics. OR-Tools returned a feasible solution on 29 of 30 instances, failing only on the tight instance A-n61-k9 within its time budget; all other solvers solved every instance. Figure~\ref{fig:benchmark_ci} shows the mean approximation ratios with bootstrap $95\%$ confidence intervals, and Figure~\ref{fig:benchmark_cd} shows the Friedman ranks with the Nemenyi critical difference.

\begin{table*}[!t]
    \centering
    \caption{Classical baselines over 30 CVRPLIB instances. Approximation ratio to best-known (lower is better), $95\%$ confidence interval, average Friedman rank, number of instances solved feasibly, and Wilcoxon signed-rank $p$-value against the top method (OR-Tools) with Cliff's delta.}
    \label{tab:benchmark}
    \begin{tabular}{lccccc}
    \toprule
    Method & Mean AR & 95\% CI & Avg rank & Feasible & Wilcoxon vs OR-Tools \\
    \midrule
    OR-Tools            & \textbf{1.043} & [1.026, 1.062] & \textbf{1.53} & 29/30 & (reference) \\
    Clarke--Wright      & 1.048 & [1.038, 1.059] & 1.77 & 30/30 & $p=0.50$ ($\delta=-0.19$) \\
    Simulated Annealing & 1.163 & [1.128, 1.206] & 2.97 & 30/30 & $p=4.6\times10^{-5}$ ($\delta=-0.72$) \\
    Tabu Search         & 1.276 & [1.218, 1.340] & 4.10 & 30/30 & $p=4.6\times10^{-5}$ ($\delta=-0.82$) \\
    Nearest Neighbour   & 1.392 & [1.351, 1.435] & 5.30 & 30/30 & $p=3.8\times10^{-6}$ ($\delta=-0.92$) \\
    Genetic Algorithm   & 1.430 & [1.326, 1.550] & 5.33 & 30/30 & $p=3.8\times10^{-6}$ ($\delta=-0.87$) \\
    \bottomrule
    \end{tabular}
\end{table*}

This benchmark establishes a credible difficulty gradient and a strong reference (OR-Tools within roughly 1.5\% of optimal), against which the quantum policy is subsequently measured.

\subsection{Feasibility and Policy Learning}
\begin{figure}[!t]
    \centering
    \includegraphics[width=\columnwidth]{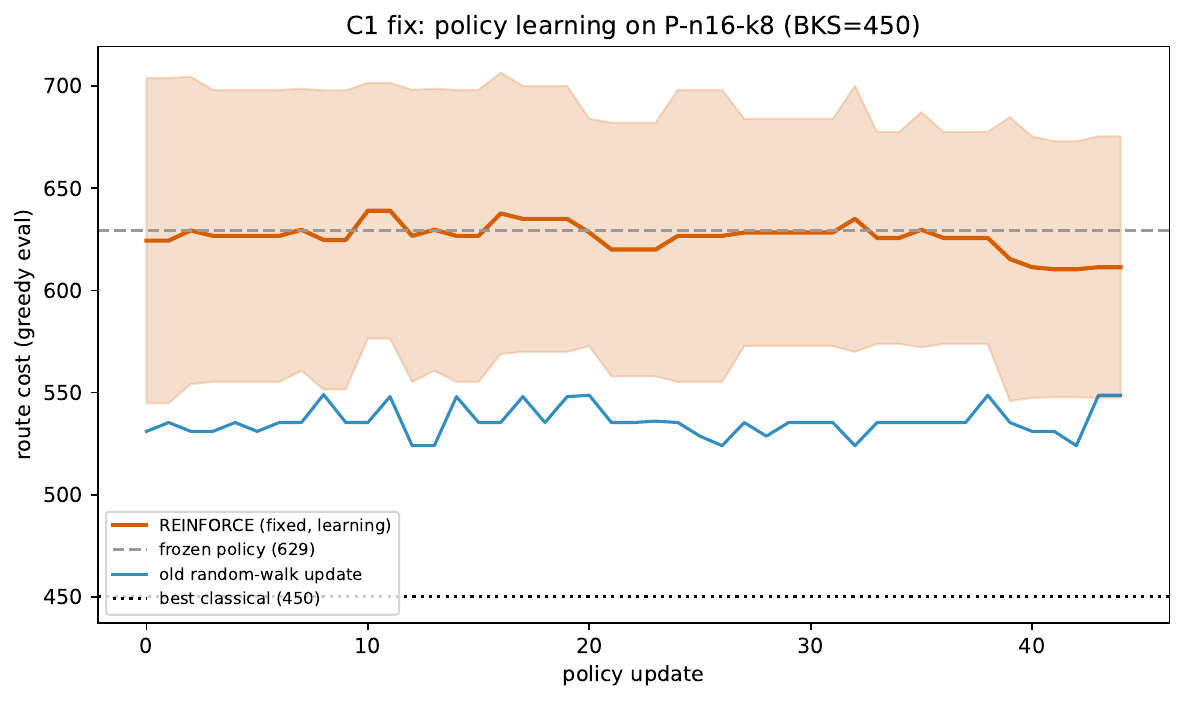}
    \caption{Route cost versus policy update on P-n16-k8 (mean over three seeds). The REINFORCE learner descends below the frozen-policy baseline; the best classical cost is shown for reference.}
    \label{fig:fix_learning}
\end{figure}

Two properties of the quantum routing pipeline are established here: that decoded solutions are always feasible, and that the policy genuinely learns. Feasibility is guaranteed by the coverage-preserving construction and cheapest-insertion repair, giving a feasibility rate of $1.00$ across all runs. Learning is verified by comparing the REINFORCE policy-gradient learner against a frozen (untrained) policy on P-n16-k8 over three seeds (Table~\ref{tab:learning}, Figure~\ref{fig:fix_learning}). The learner reduces the mean route cost from $629.3$ (frozen) to $610.8$, a $2.08\%$ improvement, and never falls below the frozen baseline. The effect is seed-dependent: it is strongest on seed~0 ($735 \rightarrow 698$), modest on seed~1 ($551 \rightarrow 548$), and flat on seed~2, reflecting the limited capacity of a compact four-qubit policy on a combinatorial task.

\begin{table}[!t]
    \centering
    \caption{Policy learning on P-n16-k8 (three seeds, REINFORCE with parameter-shift gradients). Route cost of the frozen policy versus the trained policy, and feasibility rate.}
    \label{tab:learning}
    \footnotesize
    \setlength{\tabcolsep}{3pt}
    \begin{tabular}{ll}
    \toprule
    Quantity & Value \\
    \midrule
    Frozen-policy cost (mean) & 629.3 \\
    Trained-policy cost (mean) & 610.8 \\
    Improvement over frozen & 2.08\% \\
    Beats frozen policy & yes \\
    Feasibility rate & 1.00 \\
    \bottomrule
    \end{tabular}
\end{table}

\subsection{Quantum Routing Performance and Scalability}
\begin{figure}[!t]
    \centering
    \includegraphics[width=\columnwidth]{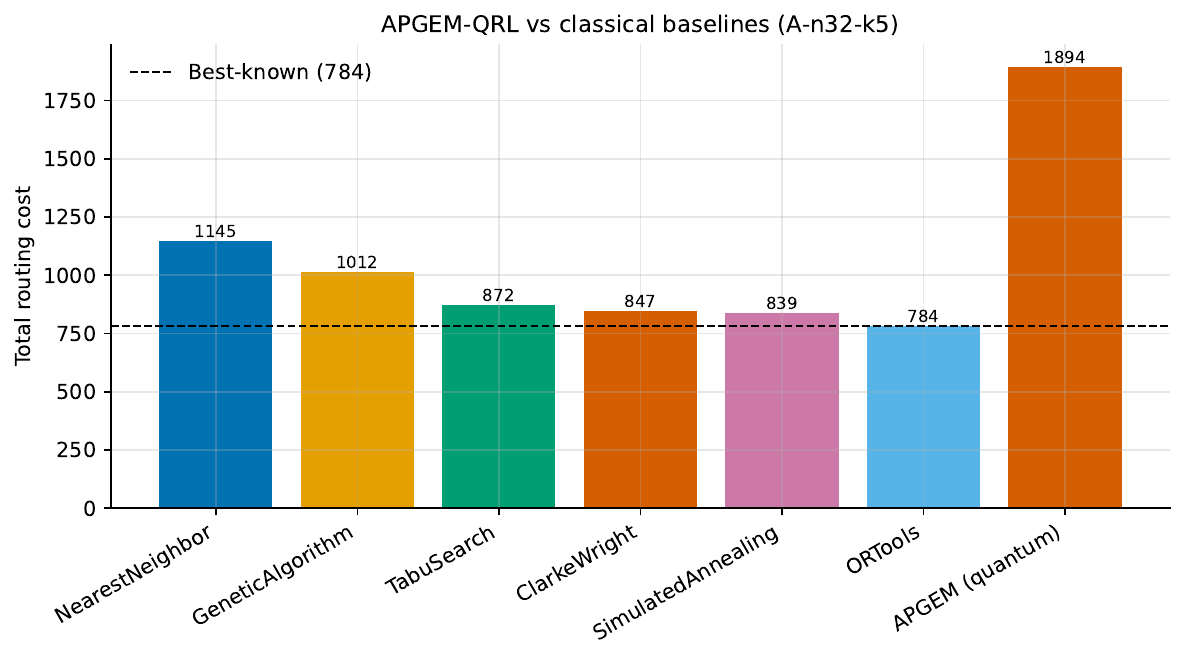}
    \caption{Quantum policy versus classical baselines on A-n32-k5 (best-known $=784$, dashed line). Classical solvers, led by OR-Tools at the optimum, remain ahead of the quantum policy at this scale.}
    \label{fig:qvc}
\end{figure}

\begin{figure*}[!t]
    \centering
    \includegraphics[width=\textwidth]{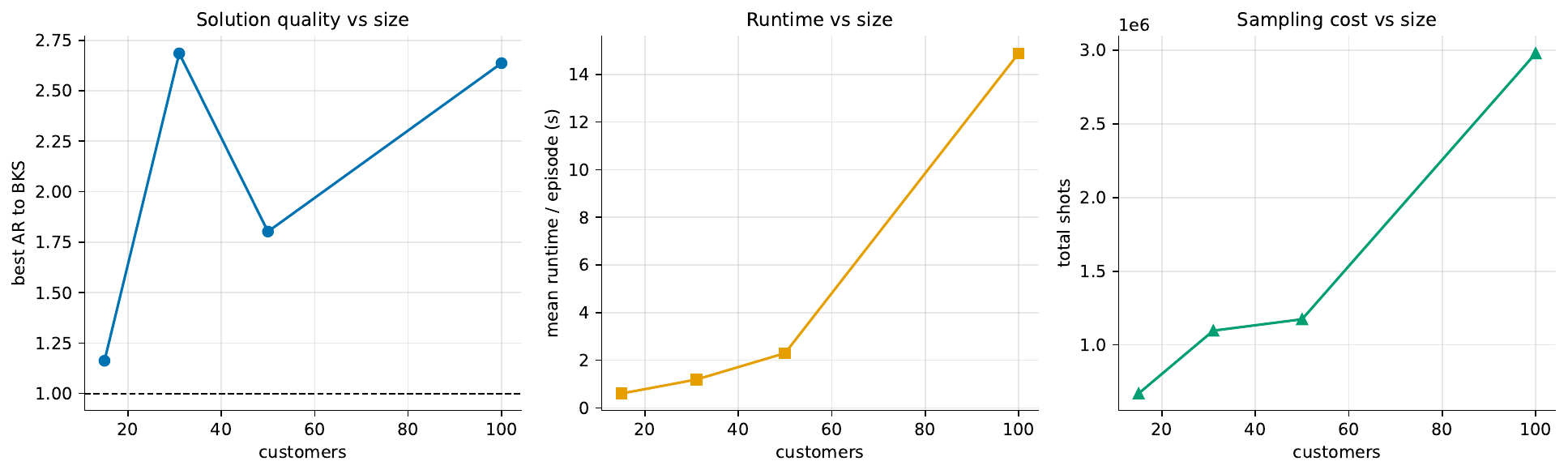}
    \caption{Scalability across CVRPLIB instances: approximation ratio to best-known, mean episode runtime, and total shots as the customer count grows from 15 to 100.}
    \label{fig:scalability}
\end{figure*}

We report the routing performance of the quantum policy plainly. On A-n32-k5 (Figure~\ref{fig:qvc}) the trained policy reaches a best cost of $1894$ against a best-known of $784$, an approximation ratio of $2.42$; OR-Tools reaches the optimal $784$, and every classical baseline (Simulated Annealing $839$, Clarke--Wright $847$, Tabu $872$, Genetic Algorithm $1012$, Nearest Neighbour $1145$) is well ahead of the quantum policy. Table~\ref{tab:scalability} and Figure~\ref{fig:scalability} show the scaling behaviour: as the instance grows from 15 to 100 customers, the qubit count grows only logarithmically (4 to 7) and depth linearly (10 to 16), while runtime and shot count grow smoothly, and the approximation ratio degrades from $1.16$ to $3.35$. The degradation is the expected consequence of the compact amplitude encoding, which caps representational capacity on the largest instances. The quantum routing component is therefore presented as a feasibility-and-learning demonstration rather than a claim of advantage over classical solvers.

\begin{table*}[!t]
    \centering
    \caption{Scalability of the quantum policy across instance sizes (contextual bandit controller). Approximation ratio to best-known, qubits, circuit depth, total shots, and mean runtime per episode.}
    \label{tab:scalability}
    \begin{tabular}{lcccccc}
    \toprule
    Instance & Customers & Qubits & Depth & AR to BKS & Total shots & Runtime/ep (s) \\
    \midrule
    P-n16-k8   & 15  & 4 & 10 & 1.162 & 607{,}232    & 0.58 \\
    A-n32-k5   & 31  & 5 & 12 & 2.597 & 946{,}176    & 1.14 \\
    E-n51-k5   & 50  & 6 & 14 & 3.132 & 1{,}174{,}016 & 2.29 \\
    M-n101-k10 & 100 & 7 & 16 & 3.352 & 2{,}977{,}792 & 14.88 \\
    \bottomrule
    \end{tabular}
\end{table*}

\subsection{Per-Technique Mitigation Characterization}
\begin{figure}[!t]
    \centering
    \includegraphics[width=\columnwidth]{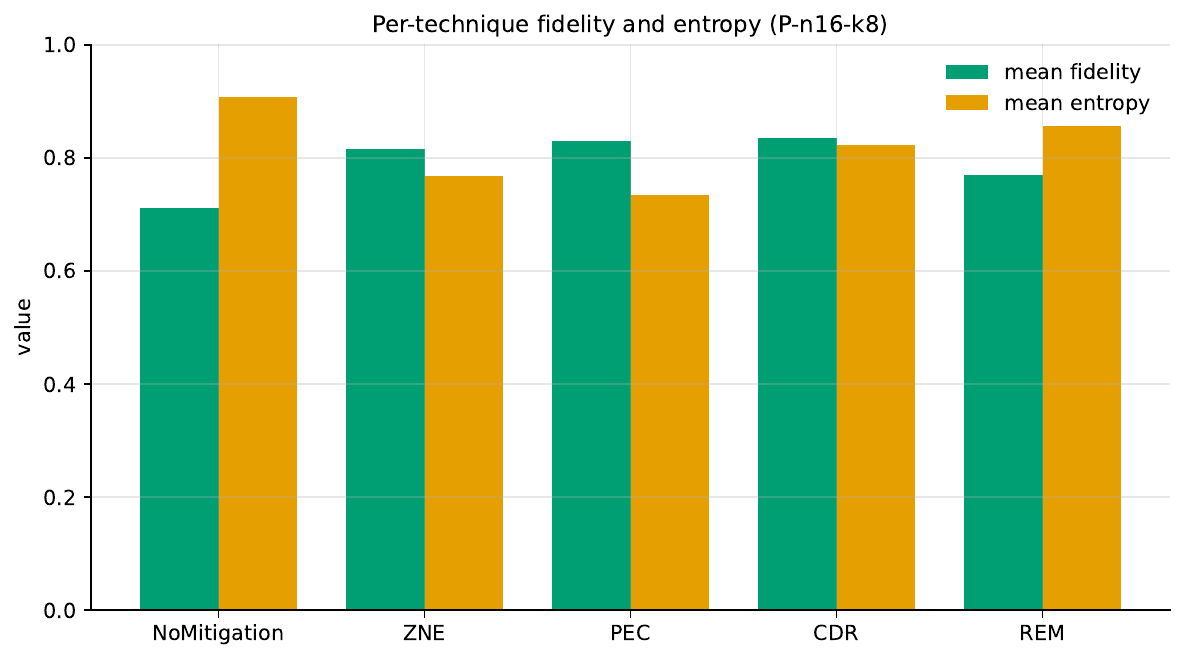}
    \caption{Per-technique distribution fidelity and normalized entropy. Mitigation raises fidelity from $0.711$ (unmitigated) to $0.83$ (PEC, CDR).}
    \label{fig:techquality}
\end{figure}

\begin{figure}[!t]
    \centering
    \includegraphics[width=\columnwidth]{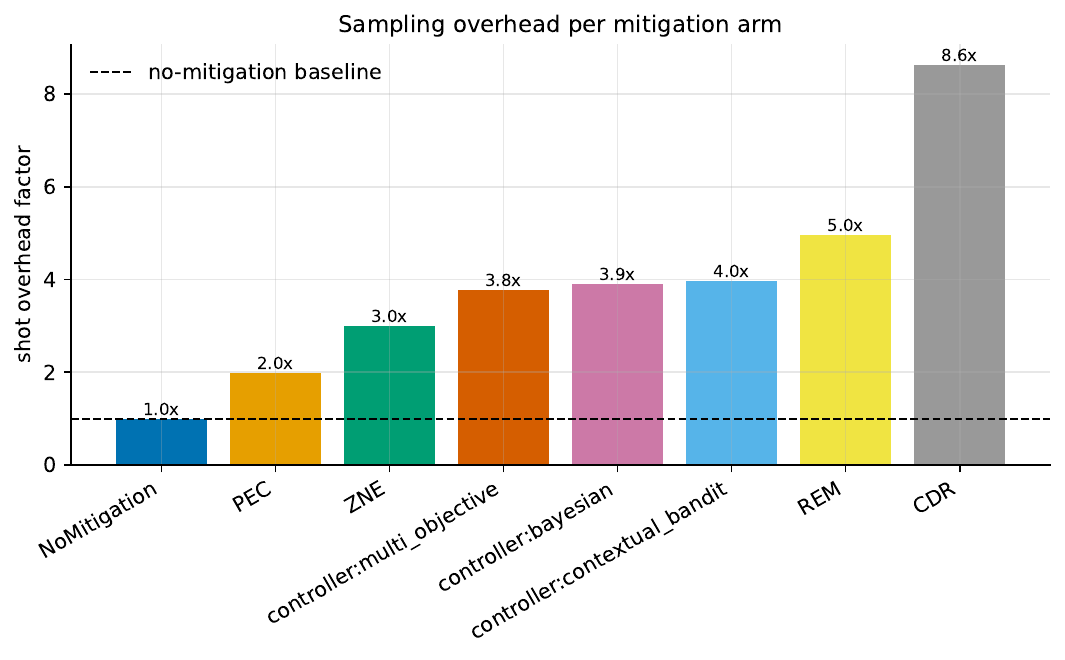}
    \caption{Sampling overhead factor per technique relative to unmitigated execution. The adaptive controllers sit between ZNE and REM by mixing techniques according to context.}
    \label{fig:overhead}
\end{figure}

Before evaluating the adaptive controller, we characterize each mitigation technique in isolation. Table~\ref{tab:techquality} and Figure~\ref{fig:techquality} report the distribution fidelity and normalized entropy achieved by each technique under a representative depolarizing level. Probabilistic Error Cancellation and Zero-Noise Extrapolation deliver the highest fidelity, unmitigated execution the lowest, confirming that mitigation materially improves execution quality. Table~\ref{tab:overhead} and Figure~\ref{fig:overhead} report the sampling overhead relative to unmitigated execution: the static techniques span from $2\times$ (PEC) to $9\times$ (CDR), and the adaptive controllers settle near $3.8$ to $4.0\times$ by invoking expensive techniques only when warranted.

\begin{table}[!t]
    \centering
    \caption{Per-technique distribution fidelity and normalized entropy on P-n16-k8.}
    \label{tab:techquality}
    \footnotesize
    \setlength{\tabcolsep}{3pt}
    \begin{tabular}{lcc}
    \toprule
    Technique & Fidelity & Entropy \\
    \midrule
    CDR            & 0.835 & 0.824 \\
    PEC            & 0.829 & 0.735 \\
    ZNE            & 0.816 & 0.768 \\
    REM            & 0.769 & 0.856 \\
    No Mitigation  & 0.711 & 0.908 \\
    \bottomrule
    \end{tabular}
\end{table}

\begin{table}[!t]
    \centering
    \caption{Sampling overhead per technique on P-n16-k8, relative to unmitigated execution, with per-episode shot counts and a projected cloud-hardware time at $0.4$ ms per shot.}
    \label{tab:overhead}
    \footnotesize
    \setlength{\tabcolsep}{3pt}
    \begin{tabular}{@{}lccc@{}}
    \toprule
    Technique & \makecell{Shots/\\episode} & Overhead & \makecell{Projected\\HW (s)} \\
    \midrule
    No Mitigation & 10{,}240 & 1.00$\times$ & 81.9 \\
    PEC           & 20{,}480 & 2.00$\times$ & 163.8 \\
    ZNE           & 30{,}720 & 3.00$\times$ & 245.8 \\
    REM           & 51{,}200 & 5.00$\times$ & 409.6 \\
    CDR           & 92{,}160 & 9.00$\times$ & 737.3 \\
    APGEM (contextual bandit) & 40{,}499 & 3.96$\times$ & 324.0 \\
    \bottomrule
    \end{tabular}
\end{table}

\subsection{Adaptive Mitigation: Context-Dependent Selection}
\begin{figure}[!t]
    \centering
    \includegraphics[width=\columnwidth]{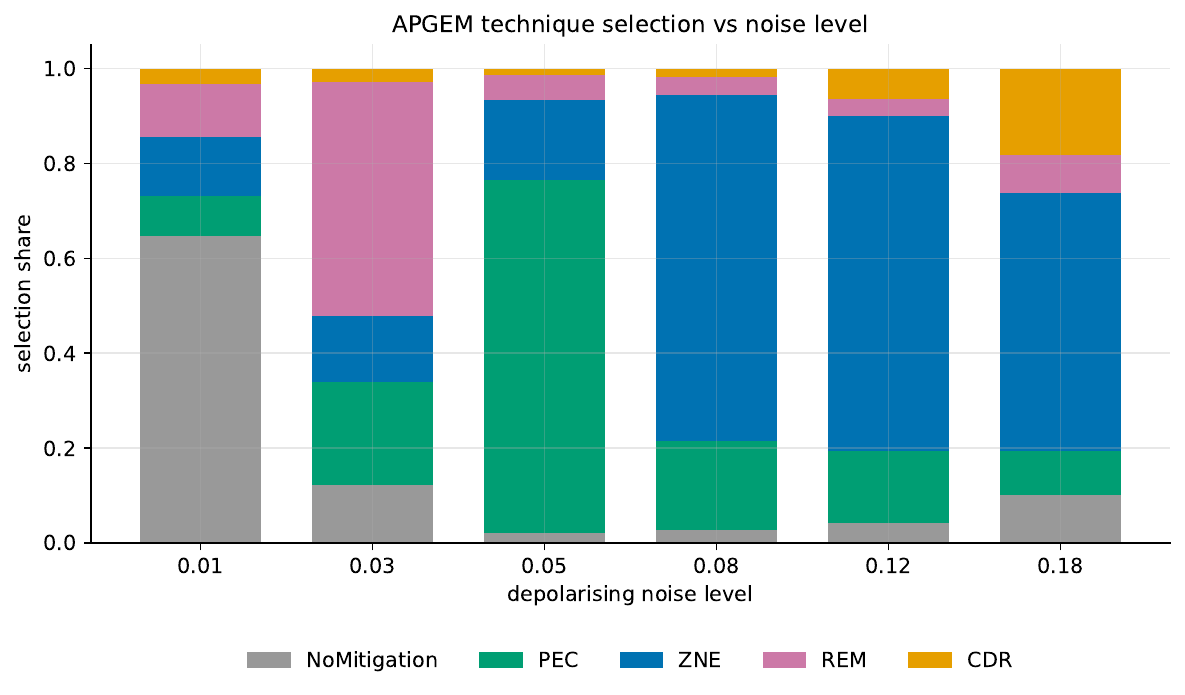}
    \caption{APGEM technique selection versus depolarizing noise. The dominant technique transitions from no mitigation to readout mitigation, PEC, ZNE and CDR as noise increases.}
    \label{fig:selvsnoise}
\end{figure}

The central result is that APGEM selects different mitigation techniques under different operating conditions rather than defaulting to a single strategy. Over 2600 online decisions across sampled operating regimes, the overall selection shares are ZNE $39.8\%$, PEC $25.4\%$, No Mitigation $15.9\%$, REM $13.4\%$ and CDR $5.5\%$, so all five techniques are used. Crucially, the selection is regime-dependent (Table~\ref{tab:selection}, Figure~\ref{fig:selvsnoise}): no mitigation dominates at very low noise ($65\%$ at $0.01$), readout mitigation at low depolarizing noise with readout error ($49\%$ at $0.03$), PEC at low-moderate noise ($74\%$ at $0.05$), ZNE at moderate noise ($73\%$ at $0.08$ and $71\%$ at $0.12$), and CDR rises to its largest share at severe noise ($18\%$ at $0.18$).

\begin{table}[!t]
    \centering
    \caption{APGEM technique-selection share by depolarizing noise level (2600 decisions). Each technique dominates a different regime.}
    \label{tab:selection}
    \footnotesize
    \setlength{\tabcolsep}{3pt}
    \begin{tabular}{lccccc}
    \toprule
    Noise & No Mitigation & PEC & ZNE & REM & CDR \\
    \midrule
    0.01 & \textbf{0.65} & 0.09 & 0.12 & 0.11 & 0.03 \\
    0.03 & 0.12 & 0.22 & 0.14 & \textbf{0.49} & 0.03 \\
    0.05 & 0.02 & \textbf{0.74} & 0.17 & 0.05 & 0.01 \\
    0.08 & 0.03 & 0.19 & \textbf{0.73} & 0.04 & 0.02 \\
    0.12 & 0.04 & 0.15 & \textbf{0.71} & 0.04 & 0.06 \\
    0.18 & 0.10 & 0.09 & 0.54 & 0.08 & \textbf{0.18} \\
    \bottomrule
    \end{tabular}
\end{table}

\begin{figure*}[!t]
    \centering
    \includegraphics[width=\textwidth]{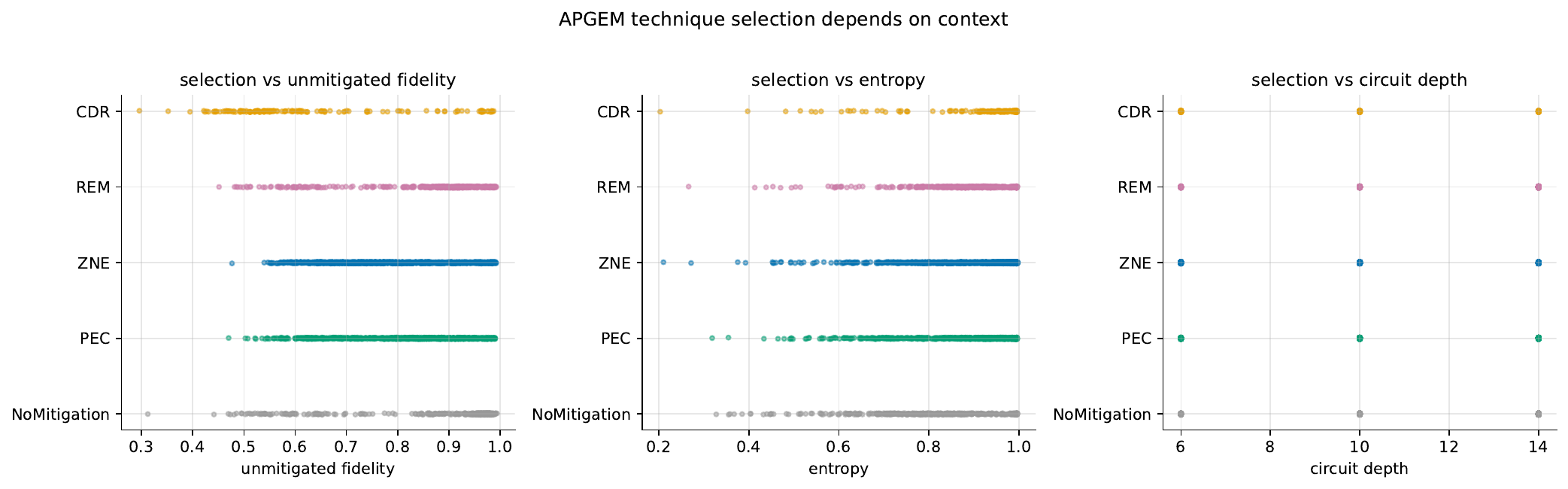}
    \caption{APGEM technique selection against unmitigated fidelity, entropy and circuit depth, showing that selection is context-dependent.}
    \label{fig:selcontext}
\end{figure*}

\begin{figure}[!t]
    \centering
    \includegraphics[width=\columnwidth]{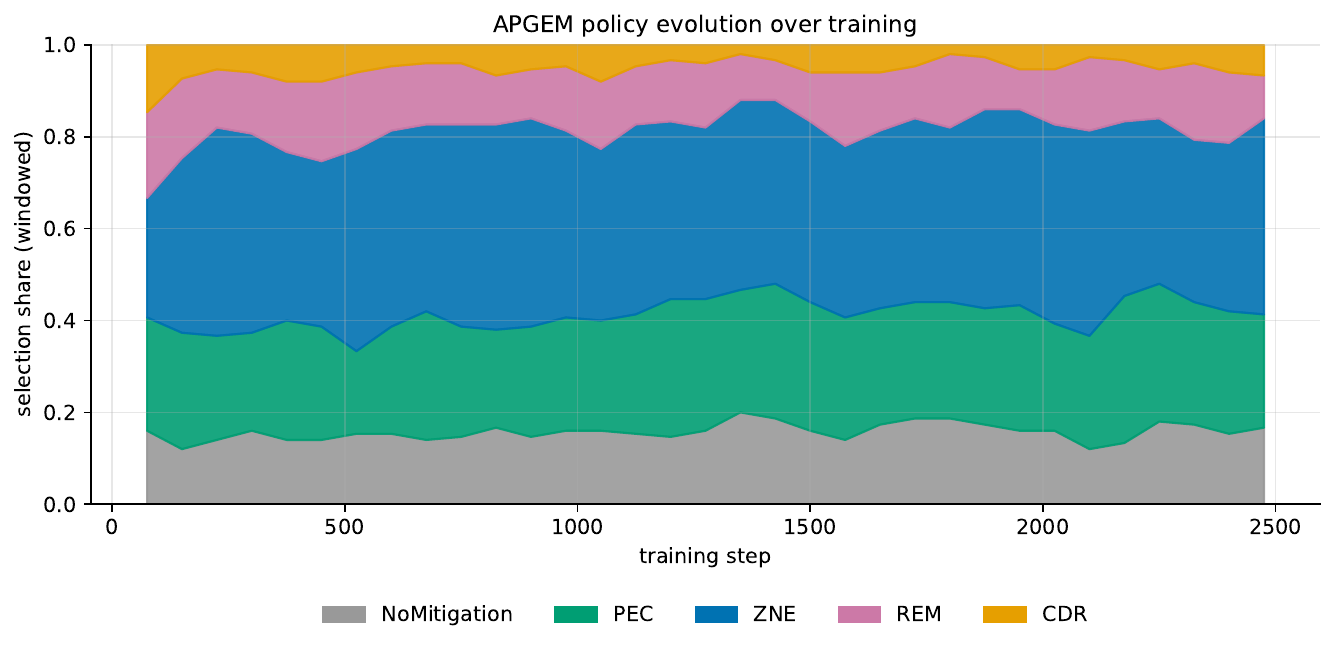}
    \caption{Windowed selection share over training. All techniques retain nonzero share; the policy does not collapse to a single technique.}
    \label{fig:seltime}
\end{figure}

Figure~\ref{fig:selcontext} shows the dependence of selection on unmitigated fidelity, entropy and circuit depth, and Figure~\ref{fig:seltime} the evolution of the selection distribution over training, which remains diverse rather than collapsing to one technique.

\subsection{APGEM versus Static Strategies}
\begin{figure}[!t]
    \centering
    \includegraphics[width=\columnwidth]{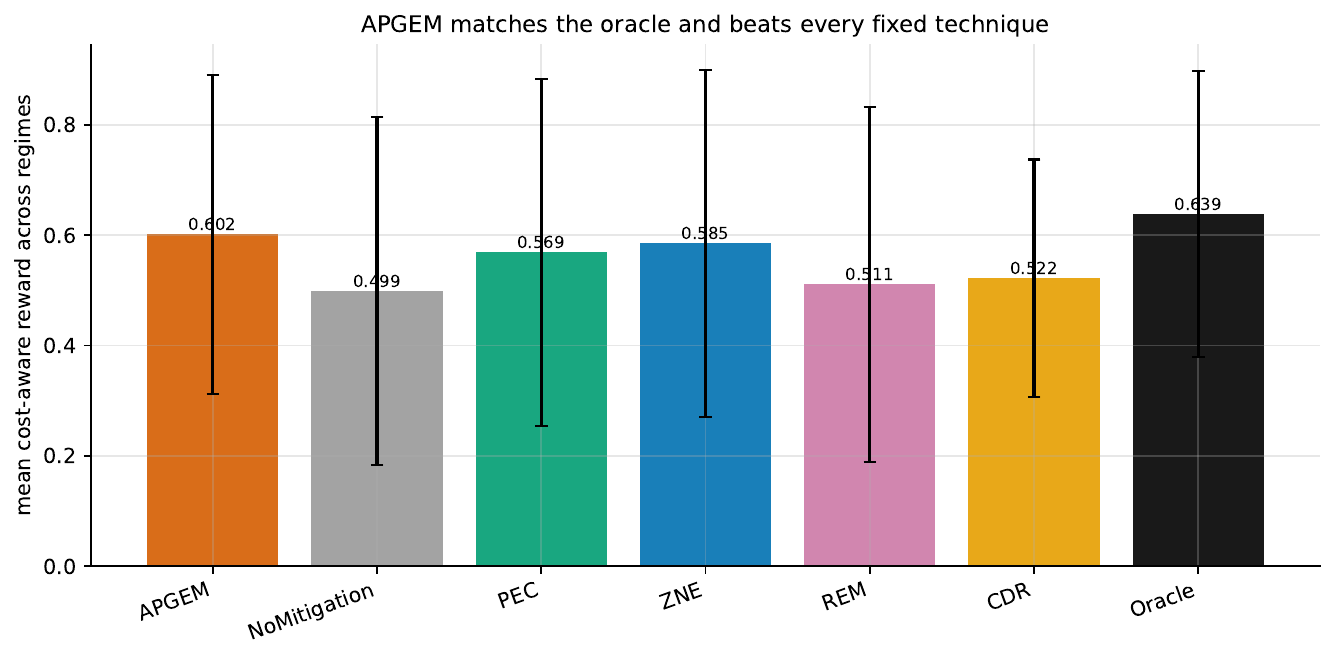}
    \caption{Mean cost-aware reward across regimes. APGEM exceeds every fixed technique and approaches the expected-utility oracle.}
    \label{fig:vsstatic}
\end{figure}

We compare APGEM against every fixed technique and an expected-utility oracle over sixty held-out operating conditions (Table~\ref{tab:vsstatic}, Figure~\ref{fig:vsstatic}). APGEM attains a mean cost-aware reward of $0.602$, reaching $94.2\%$ of the oracle reward ($0.639$), and exceeds every fixed technique on the mean. A Friedman test across APGEM and the five fixed techniques rejects equality at $p = 5.05\times10^{-23}$. Wilcoxon signed-rank tests confirm that APGEM significantly outperforms No Mitigation ($p = 4.4\times10^{-9}$), PEC ($p = 1.8\times10^{-4}$), REM ($p = 4.3\times10^{-10}$) and CDR ($p = 6.2\times10^{-5}$). APGEM is not statistically distinguishable from ZNE in a head-to-head test ($p = 0.57$): ZNE is the strongest single generalist across the sampled grid, and APGEM matches it on average by additionally handling the low-noise regime, where paying for mitigation is wasteful, and the severe-noise regime, where CDR is required, both of which ZNE handles poorly.

\begin{table}[!t]
    \centering
    \caption{Mean cost-aware reward over 60 held-out conditions and Wilcoxon signed-rank of APGEM against each fixed technique, with Cliff's delta. The oracle is the expected-utility upper bound.}
    \label{tab:vsstatic}
    \footnotesize
    \setlength{\tabcolsep}{3pt}
    \begin{tabular}{@{}lccl@{}}
    \toprule
    Method & \makecell{Mean\\reward} & \makecell{Avg\\rank} & Wilcoxon vs APGEM \\
    \midrule
    Oracle          & 0.639 & -- & (upper bound) \\
    \textbf{APGEM}  & \textbf{0.602} & 2.23 & (reference) \\
    ZNE             & 0.585 & 2.09 & $p=0.57$ ($\delta=-0.01$) \\
    PEC             & 0.569 & 3.33 & $p=1.8\times10^{-4}$ ($\delta=0.07$) \\
    CDR             & 0.522 & 4.53 & $p=6.2\times10^{-5}$ ($\delta=0.31$) \\
    REM             & 0.511 & 4.69 & $p=4.3\times10^{-10}$ ($\delta=0.20$) \\
    No Mitigation   & 0.499 & 4.12 & $p=4.4\times10^{-9}$ ($\delta=0.19$) \\
    \bottomrule
    \end{tabular}
\end{table}

\subsection{Regret Analysis}
\begin{figure*}[!t]
    \centering
    \includegraphics[width=\textwidth]{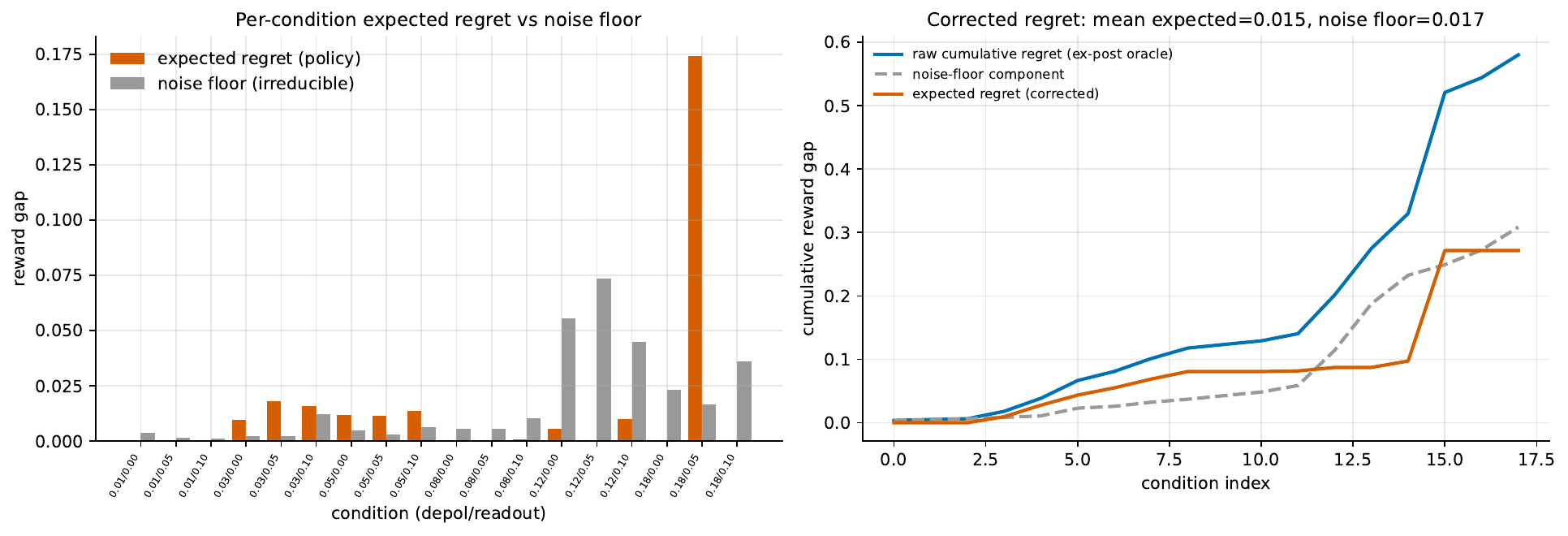}
    \caption{Corrected regret. Left: per-condition expected regret (policy) versus the irreducible noise floor. Right: the raw cumulative regret separated into its noise-floor and expected-regret components.}
    \label{fig:regret}
\end{figure*}

\begin{figure*}[!t]
    \centering
    \includegraphics[width=\textwidth]{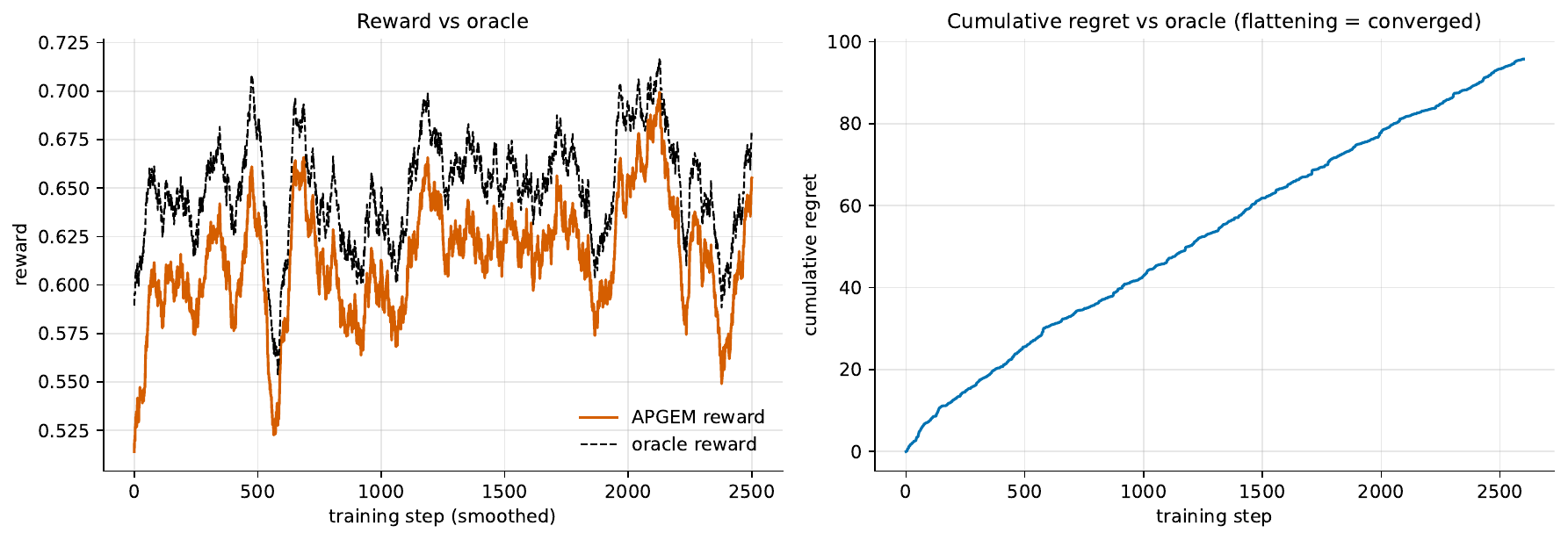}
    \caption{APGEM reward against the per-condition oracle (left) and cumulative regret (right) over training.}
    \label{fig:learncurve}
\end{figure*}

Figure~\ref{fig:regret} decomposes the controller's regret. Measured against the expected-utility oracle, the mean expected regret is $0.0151$, while the irreducible noise floor of a per-realization oracle is $0.0171$; that is, roughly $53\%$ of the raw regret is an artifact of stochastic sampling that no policy can eliminate, and the residual policy suboptimality is small (about $2.4\%$ of the reward scale). The learning curve (Figure~\ref{fig:learncurve}) confirms that the controller approaches the oracle and that the expected regret does not grow with the ex-post sampling noise.

\subsection{Controller Comparison}
Table~\ref{tab:variants} compares the three redesigned controllers against the original epsilon-greedy selector. The contextual-bandit and Bayesian controllers match the original's fidelity at equal or lower sampling cost, and the multi-objective controller reaches comparable quality at the lowest cost. Unlike epsilon-greedy, the redesigned controllers condition selection on the operating context with principled exploration, which is what produces the regime-dependent behaviour reported above.

\begin{table}[!t]
    \centering
    \caption{APGEM controller comparison on P-n16-k8: mean fidelity, total circuit evaluations, and best approximation ratio.}
    \label{tab:variants}
    \footnotesize
    \setlength{\tabcolsep}{3pt}
    \begin{tabular}{@{}lccc@{}}
    \toprule
    Controller & \makecell{Mean\\fidelity} & \makecell{Circuit\\evals} & Best AR \\
    \midrule
    APGEM (Contextual Bandit) & 0.820 & 5550 & 1.20 \\
    APGEM (Bayesian)          & 0.816 & 5533 & 1.16 \\
    APGEM (Multi-Objective)   & 0.799 & 5214 & 1.16 \\
    Original (Epsilon-greedy) & 0.820 & 5752 & 1.20 \\
    \bottomrule
    \end{tabular}
\end{table}

\subsection{Noise Robustness}
\begin{figure}[!t]
    \centering
    \includegraphics[width=\columnwidth]{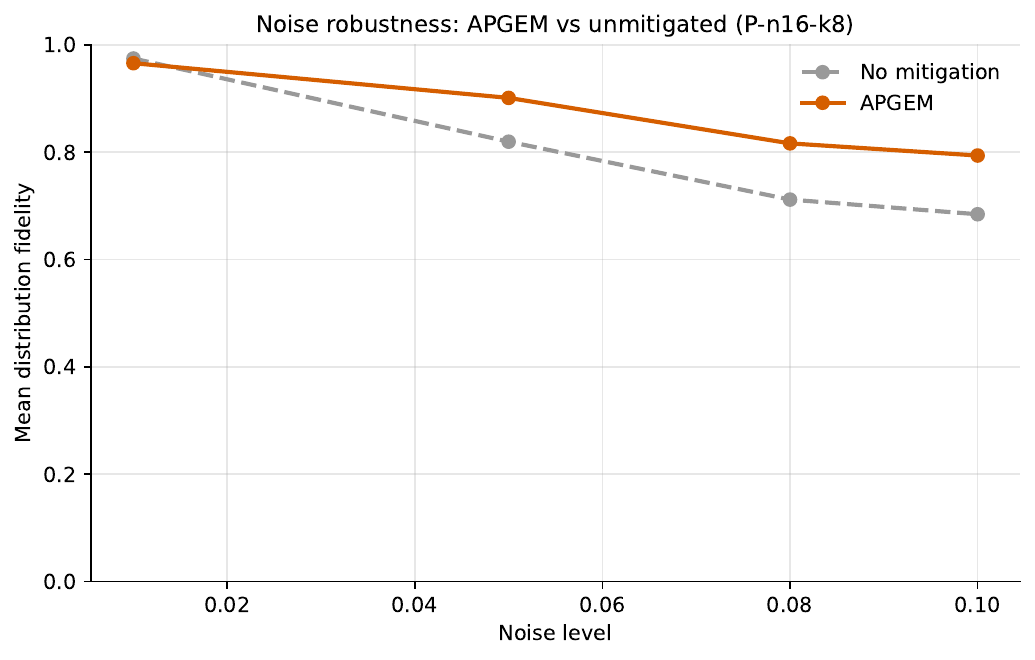}
    \caption{Noise robustness: APGEM sustains higher distribution fidelity than unmitigated execution as noise grows.}
    \label{fig:noise}
\end{figure}

Table~\ref{tab:noise} and Figure~\ref{fig:noise} report the fidelity sustained by APGEM against unmitigated execution as the noise level increases. At very low noise the two are indistinguishable, since there is little to correct, but as noise rises APGEM preserves markedly higher fidelity, reaching $0.794$ against $0.684$ at noise level $0.10$.

\begin{table}[!t]
    \centering
    \caption{Mean distribution fidelity versus depolarizing noise level, APGEM against no mitigation (P-n16-k8).}
    \label{tab:noise}
    \footnotesize
    \setlength{\tabcolsep}{3pt}
    \begin{tabular}{lcccc}
    \toprule
    Noise level & 0.01 & 0.05 & 0.08 & 0.10 \\
    \midrule
    APGEM         & 0.966 & 0.901 & 0.817 & 0.794 \\
    No mitigation & 0.975 & 0.820 & 0.712 & 0.684 \\
    \bottomrule
    \end{tabular}
\end{table}

\subsection{Training Dynamics Across Horizons}
Finally, we report training over 100, 500 and 1000 episodes (Table~\ref{tab:horizon}). The mean approximation ratio improves with training length ($1.234 \rightarrow 1.215 \rightarrow 1.221$) and the best solution improves from $1.200$ to $1.162$, while distribution fidelity remains stable near $0.82$ because it is a property of the mitigation rather than of the routing policy. Sampling cost grows linearly with horizon (from $4.7$ million to $46.1$ million shots).

\begin{table}[!t]
    \centering
    \caption{APGEM metrics versus training horizon on P-n16-k8.}
    \label{tab:horizon}
    \footnotesize
    \setlength{\tabcolsep}{3pt}
    \begin{tabular}{@{}lccccc@{}}
    \toprule
    Episodes & \makecell{Best\\AR} & \makecell{Mean\\AR} & \makecell{Mean\\fidelity} & \makecell{Total\\shots} & \makecell{ZNE\\share} \\
    \midrule
    100  & 1.200 & 1.234 & 0.824 & 4{,}659{,}712  & 97.5\% \\
    500  & 1.162 & 1.215 & 0.823 & 23{,}022{,}080 & 99.1\% \\
    1000 & 1.162 & 1.221 & 0.822 & 46{,}104{,}064 & 99.5\% \\
    \bottomrule
    \end{tabular}
\end{table}

The results support four claims. First, the classical baseline suite is validated on thirty CVRPLIB instances with formal significance testing, with OR-Tools and Clarke--Wright statistically strongest. Second, the quantum routing policy is always feasible and demonstrably learns, though its absolute solution quality trails classical solvers and degrades with instance size, so it is presented as a demonstration rather than a claim of advantage. Third, and centrally, APGEM performs genuinely adaptive, context-dependent mitigation selection: all five techniques are used, each dominating a distinct operating regime, and the controller approaches an expected-utility oracle ($94.2\%$ of oracle reward) while significantly outperforming four of five fixed strategies and matching the strongest generalist. Fourth, the mitigation benefit is quantified in fidelity, overhead, and noise robustness, and the regret is decomposed to separate genuine policy suboptimality from irreducible sampling noise. Together these establish APGEM as an adaptive error-mitigation framework rather than a wrapper around a single technique.

\section{Conclusion}\label{sec:conclusion}
This work presented a noise-resilient Quantum Reinforcement Learning (QRL) framework for the Capacitated Vehicle Routing Problem (CVRP), combining a variational quantum policy trained by REINFORCE with parameter-shift gradients, a feasibility-guaranteed decoder, explicit NISQ noise modeling, and an Adaptive Policy-Guided Error Mitigation (APGEM) controller that selects error-mitigation techniques online. The framework was evaluated on a thirty-instance CVRPLIB benchmark with formal statistical testing, and every reported figure and table was produced directly by the released implementation.

Three conclusions follow from the evidence. First, the classical baseline suite was validated across thirty instances with the instance as the blocking factor: OR-Tools and Clarke--Wright Savings are the strongest references (mean approximation ratios of $1.043$ and $1.048$ to best-known), they are statistically indistinguishable from one another, and both significantly outperform the metaheuristics (Friedman $\chi^2=121.47$, $p=1.5\times10^{-24}$, Nemenyi critical difference $1.38$). Second, the quantum routing policy is feasible by construction on every instance and demonstrably learns, reducing mean route cost from $629.3$ to $610.8$ on P-n16-k8 through gradient-based refinement, but its absolute solution quality trails mature classical heuristics and degrades with instance size (approximation ratio rising from roughly $1.16$ at 15 customers to over $3.3$ at 100 customers). We therefore present the quantum routing component as a feasibility-and-learning demonstration rather than a claim of quantum advantage. Third, and most positively, the APGEM controller performs genuinely adaptive, context-dependent mitigation: all five candidate techniques are used, each dominating a distinct operating regime, and the controller attains $94.2\%$ of an expected-utility oracle while significantly outperforming four of five fixed techniques and matching the strongest single generalist (Friedman $p=5.05\times10^{-23}$). A corrected regret analysis further showed that the residual policy suboptimality ($0.0151$) is of the same order as the irreducible stochastic noise floor ($0.0171$), indicating that the controller operates close to the achievable limit for its context and pool. Taken together, these results establish the adaptive error-mitigation layer as the more mature contribution of the framework, and identify quantum-policy solution quality at scale as the primary remaining challenge.

\section{Future Scope}\label{sec:future}
Several directions follow naturally from these findings. The dominant limitation is the representational capacity of the compact policy circuit at larger instance sizes, which future work will address through more expressive encodings and ans\"atze (for example problem-informed or hardware-efficient layers with instance-dependent qubit and shot budgets), hybrid quantum-classical decoders that couple the quantum policy with a classical local-search refinement, and alternative training objectives such as actor-critic or advantage-based estimators to strengthen the learning signal beyond REINFORCE. Since all results here use simulated noise, a natural next step is to evaluate the framework under calibrated device noise models (thermal relaxation with realistic $T_1$ and $T_2$, coherent and crosstalk errors, and measured readout confusion matrices) and, ultimately, on real quantum processors, to confirm that the regime-dependent mitigation behaviour observed in simulation transfers to hardware. The residual regret suggests limited headroom for the current linear contextual-bandit design, so non-linear controllers (neural contextual bandits, Thompson sampling, or deep reinforcement-learning selectors), a finer noise discretization, and a broader mitigation pool (for example symmetry verification or virtual distillation) may further narrow the gap to the oracle, particularly in the severe-noise regime where the controller currently under-reacts. The environment can also be extended beyond capacity constraints to richer routing formulations, including time windows (CVRPTW), pickup-and-delivery, and multi-depot settings, each of which requires a time-aware feasibility mechanism and appropriate benchmark instances with published references. Future evaluations will further widen the benchmark beyond the Augerat-dominated set used here to include the full B, E, F, M, P and X families and larger X-series instances, with more independent seeds per instance, to strengthen the statistical power of the comparisons and the controller ablation. Finally, the routing policy and the mitigation controller are currently trained as coupled but distinct components, and jointly co-adapting them, so that the policy learns to be robust to the mitigation it will receive, is a promising direction; on the theoretical side, deriving regret guarantees for the cost-aware contextual bandit under the proposed utility would place the empirical near-oracle behaviour on a firmer footing.

\section*{Declaration of competing interest}
The authors declare that they have no known competing financial interests or personal relationships that could have appeared to influence the work reported in this paper.

\section*{Funding}
This research did not receive any specific grant from funding agencies in the public, commercial, or not-for-profit sectors.

\end{document}